\documentclass[11pt]{article}

\usepackage[final]{acl}
\usepackage{alltt}
\usepackage{amssymb}     
\usepackage{fancyvrb}    
\usepackage{times}
\usepackage{latexsym}
\usepackage[dvipsnames]{xcolor}
\usepackage[most]{tcolorbox}
\usepackage{soul}
\usepackage{fvextra}
\usepackage{url}

\usepackage{pifont}
\tcbset{
  promptstyle/.style={
    colback=gray!15,  
    colframe=gray!60,  
    boxrule=0.4pt,     
    arc=2pt,           
    left=4pt,right=4pt,top=4pt,bottom=4pt, 
    fonttitle=\bfseries,
  }
}
\newtcolorbox{promptblock}[1][]{promptstyle,#1}
\usepackage{colortbl}
\usepackage[T1]{fontenc}
\usepackage{arydshln}
\usepackage[utf8]{inputenc}

\usepackage{microtype}
\usepackage{subcaption}  
\usepackage{placeins}
\usepackage{inconsolata}
\usepackage{xcolor}      
\usepackage{graphicx}
\usepackage{booktabs}
\usepackage{multicol}
\usepackage{multirow}
\usepackage{xspace}
\usepackage{array}
\newcolumntype{C}{>{\centering\arraybackslash}p{0.055\textwidth}}
\newcolumntype{D}{>{\centering\arraybackslash}p{0.03\textwidth}}
\newcolumntype{E}{>{\centering\arraybackslash}p{0.04\textwidth}}
\newcommand{\modelname}{UI-Copilot-7B\xspace}
\newcommand{\methodname}{TIPO\xspace}

\makeatletter
\newcommand{\thickhline}{%
    \noalign {\ifnum 0=`}\fi \hrule height 1pt
    \futurelet \reserved@a \@xhline
}
\makeatother
\newcolumntype{I}{!{\vrule width 1pt}}
\newcolumntype{i}{!{\vrule width 0.6pt}}
\definecolor{androidworld}{RGB}{216, 228, 244}
\definecolor{memguibench}{RGB}{252, 228, 202}
\definecolor{memory}{RGB}{216, 228, 244}
\definecolor{other}{RGB}{234, 235, 235}
\definecolor{progress}{RGB}{232, 204, 204}
\definecolor{math}{RGB}{229, 190, 217}
\definecolor{cvprblue}{rgb}{0.21,0.49,0.74}
\definecolor{lightblue}{RGB}{242, 244, 250}
\definecolor{ourdelta}{RGB}{254, 252, 240}
\definecolor{copilotred}{RGB}{238, 68, 51}
\definecolor{toolblue}{RGB}{8, 111, 189}
\definecolor{yellow}{RGB}{255, 192, 0}
\definecolor{purple}{RGB}{216, 110, 204}
\definecolor{brown}{RGB}{127, 36, 28}
\definecolor{green}{RGB}{71, 172, 20}
\definecolor{orange}{RGB}{194,153,107}
\definecolor{codegray}{gray}{0.95}

\DeclareMathOperator*{\argmax}{arg\,max}  
\DeclareMathOperator*{\rolloutpi}{\textcolor{red}{\mu}}
\DeclareMathOperator*{\trainerpi}{\textcolor{blue}{\pi}}

\newcommand{\heatcell}[2]{%
  \multicolumn{1}{c}{%
    \raisebox{-.18\height}{%
      \colorbox[HTML]{#1}{%
        \hspace{4.5pt}%
        \rule{0pt}{1em}
        \raisebox{0.1em}{#2}
        \hspace{4.5pt}%
      }%
    }%
  }%
}
\title{UI-Copilot: Advancing Long-Horizon GUI Automation via\\ Tool-Integrated Policy Optimization}

\author{
\textbf{Zhengxi Lu}$^{1}$, \textbf{Fei Tang}$^{1}$, 
\textbf{Guangyi Liu}$^{1}$, \textbf{Kaitao Song}$^{2}$, \textbf{Xu Tan}$^{3}$, \textbf{Jin Ma}$^{3}$\\
\textbf{Wenqi Zhang}$^{1}$, 
\textbf{Weiming Lu}$^{1}$,
\textbf{Jun Xiao}$^{1}$,
\textbf{Yueting Zhuang}$^{1}$,
 \textbf{Yongliang Shen}$^{1}$\thanks{Corresponding author }\\[3pt]
  $^1$Zhejiang University \qquad$^2$Apple \qquad$^3$Tencent\\
  \texttt{\{zhengxilu, syl\}@zju.edu.cn} \\
  \begin{tabular}{@{}ll@{}}
  \end{tabular}}

\begin{document}
\maketitle
\begin{abstract}
MLLM-based GUI agents have demonstrated strong capabilities in complex user interface interaction tasks.
However, long-horizon scenarios remain challenging, as these agents are burdened with tasks beyond their intrinsic capabilities, suffering from memory degradation, progress confusion, and math hallucination.
To address these challenges, we present \textbf{UI-Copilot}, a collaborative framework where the GUI agent focuses on task execution while a lightweight copilot provides on-demand assistance for memory retrieval and numerical computation.
We introduce memory decoupling to separate persistent observations from transient execution context, and train the policy agent to selectively invoke the copilot as \texttt{Retriever} or \texttt{Calculator} based on task demands.
To enable effective tool invocation learning, we propose \underline{\textbf{T}}ool-\underline{\textbf{I}}ntegrated \underline{\textbf{P}}olicy \underline{\textbf{O}}ptimization (\textbf{TIPO}), which separately optimizes tool selection through single-turn prediction and task execution through on-policy multi-turn rollouts.
Experimental results show that UI-Copilot-7B achieves state-of-the-art performance on challenging MemGUI-Bench, outperforming strong 7B-scale GUI agents such as GUI-Owl-7B and UI-TARS-1.5-7B. Moreover, UI-Copilot-7B delivers a 17.1\% absolute improvement on AndroidWorld over the base Qwen model, highlighting UI-Copilot's strong generalization to real-world GUI tasks.
\end{abstract}

\section{Introduction}
\begin{figure}[t]
\centering
  \includegraphics[width=0.48\textwidth]{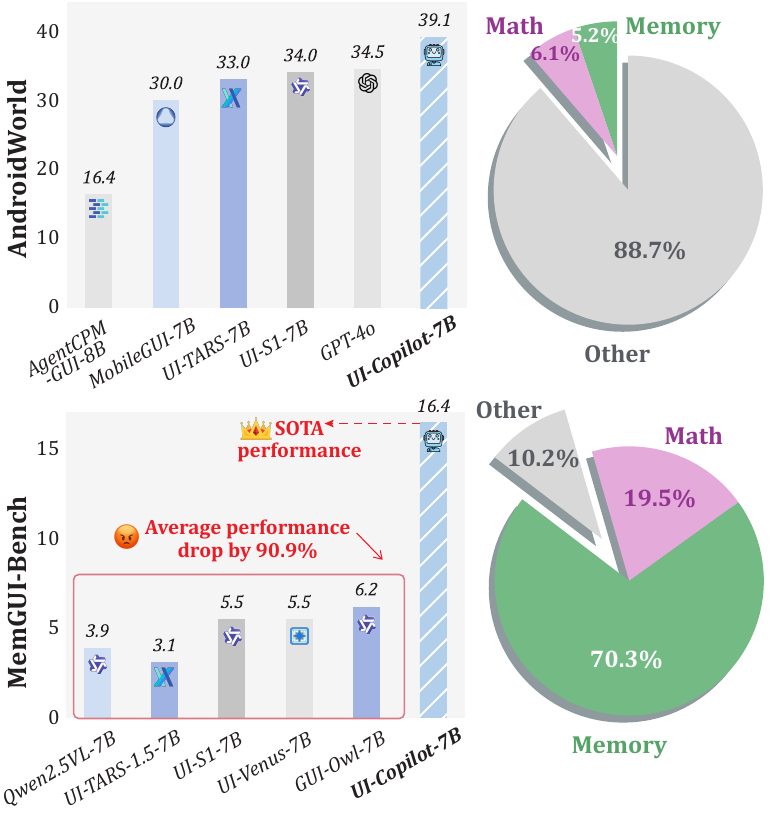}
  \caption{\textbf{Left}: Performance on dynamic GUI benchmarks. \textbf{Right}: Task distribution of these benchmarks.}
  \label{fig:benchmark_comparison}
  \vspace{-4mm}
\end{figure}
Graphical User Interface (GUI) agents are designed to interact with digital environments in a human-like manner~\citep{hu2025osagents, zhang2025appagent, wang2025opencua, tang2025survey, liu2025survey}.
Recent multimodal large language model (MLLM)-based GUI agents~\citep{ye2025mobileagentv3fundamentalagentsgui,gu2025uivenus,qin2025uitars}, trained via supervised fine-tuning (SFT) and reinforcement learning (RL), have demonstrated strong capability in solving short-horizon tasks, which typically require fewer than 10 interaction steps~\citep{rawles2024androidworld,zhao2025masbench,chen2026knowubench}, as shown in Figure~\ref{fig:benchmark_comparison} and~\ref{fig:golden_steps}).
\begin{figure*}[t]
  \includegraphics[width=\textwidth]{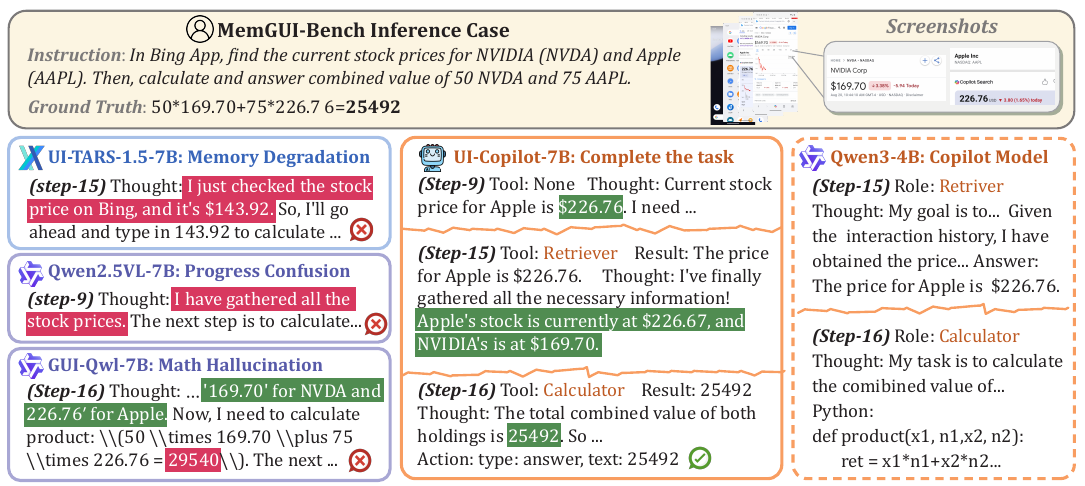}
  \caption{\textbf{MemGUI-Bench Inference Case.} Our method successfully completes the task by invoking Copilot Model, whereas other models fail due to memory degradation, progress confusion, and math hallucinations.}
  \label{fig:motivation}
  \vspace{-3mm}
\end{figure*}

However, deploying these agents in long-horizon, memory-intensive scenarios exposes fundamental limitations. As shown in Figure~\ref{fig:benchmark_comparison}, existing 7B models suffer an average performance drop of 90.90\% on MemGUI-Bench~\citep{liu2026memgui}. We identify three interconnected challenges underlying these failures (as illustrated in Figure~\ref{fig:motivation}): \textbf{1) Memory Degradation}: overloaded context causes agents to lose or misrecall critical information from earlier steps; \textbf{2) Progress Confusion}: interleaving reasoning traces with action histories obscures task state, leading to redundant actions, disordered sub-task execution, or premature termination; and \textbf{3) Math Hallucination}: numerical reasoning errors compound as incorrect intermediate results propagate through subsequent computations. 

Existing approaches address these limitations through multi-agent workflows~\citep{Agent-S2,wang2025mobileagent-e,wang2024mobileagent-2} or retrieval augmentation~\citep{liu2025learnact,li2025echotrail,xu2025raggui}. However, multi-agent workflows rely on predefined pipelines that invoke external modules regardless of actual necessity, resulting in prohibitive inference costs. Retrieval-augmented methods depend heavily on retrieval quality and fail to resolve progress confusion. We attribute these limitations to a shared underlying cause:

\vspace{-2mm}
\begin{tcolorbox}[colframe=black!50, colback=cvprblue!8, boxrule=1.5pt, arc=2mm, top=4pt, bottom=4pt, left=4pt, right=4pt,  boxsep=1pt]
\raisebox{-0.2\baselineskip}{\includegraphics[height=1\baselineskip]{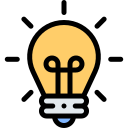}} \textit{Agents are burdened with challenges beyond its capabilities, leading to confusion under increasingly overloaded context.}
\end{tcolorbox}
\vspace{-2mm}

Our key insight is that \textit{\textbf{GUI agents should focus on task execution with lightweight context, while memory and computation are decoupled and invoked only when needed.}}
Building on this insight, we introduce \textbf{UI-Copilot}, a collaborative framework that decouples persistent observations from transient execution context. 
Detailed reasoning traces are stored locally while only concise progress summaries remain in the dialogue history, keeping the context window focused and enabling on-demand information retrieval. The policy agent selectively invokes a lightweight copilot model as \texttt{Retriever} or \texttt{Calculator}, enabling adaptive tool usage that responds to actual task demands.

To enable more effective tool invocation, we propose \textbf{T}ool-\textbf{I}ntegrated \textbf{P}olicy \textbf{O}ptimization (\textbf{TIPO}), which decouples tool prediction and task execution during training.
Tool selection is trained via single-turn supervision, while action generation learns through multi-turn rollouts conditioned on self-generated histories, aligning training dynamics with deployment conditions.
Extensive experiments demonstrate that UI-Copilot-7B achieves SOTA performance on MemGUI-Bench, where over 90\% of tasks require persistent memory, and attains 39.1\% accuracy on AndroidWorld, validating the generalization of the proposed framework. In summary, our contributions are:
\begin{itemize}
\item We propose \textbf{UI-Copilot}, a collaborative framework where the GUI agent selectively invokes a lightweight copilot for memory retrieval and numerical computation, enabling efficient long-horizon GUI navigation.
\item We introduce \textbf{memory decoupling} to separate persistent observations from transient context, effectively mitigating context overload.
\item We develop \textbf{TIPO}, a reinforcement learning algorithm that separately trains GUI agents' tool invocation and action generation.
\end{itemize}

\section{Related Work}
\subsection{Reinforcement Learning for GUI Agent}
Recent advances in multimodal models have catalyzed significant progress in GUI automation~\citep{hu2025osagents,zhang2025appagent,wang2025opencua,tang2025survey,liu2025survey,ye2025mobileagentv3fundamentalagentsgui,wu2026spark}. Inspired by DeepSeek-R1~\citep{guo2025deepseekr1}, recent work~\citep{lu2025uir1,luo2025guir1,qin2025uitars,lu2025arpogui,gu2025uivenus,tang2025guig2,du2025ttrlgui} has begun applying Group Relative Policy Optimization (GRPO)~\citep{shao2024deepseekmath} to GUI automation. However, external tool calling for GUI agent training remain un-explored.

\subsection{Memory for GUI Agent}
Memory remains a fundamental challenge for GUI agents due to the limited context windows. Recent attempts, including multi-agent workflows~\citep{wang2024mobileagent-2,wang2025mobileagent-e,Agent-S2} and few-shot Retrieval-Augmented Generation (RAG)~\citep{liu2025learnact,li2025echotrail,xu2025raggui}, aim to enhance agent performance without model fine-tuning. While effective in certain scenarios, these approaches often suffer from limited scalability and high deployment costs. Additional works incorporate history-aware training mechanisms~\citep{zhou2025hiconagent,liu2025guirise,wang2025history-aware-gui,lu2026skill0}, but still face memory degradation for memory-intensive, long-horizon GUI tasks.

\section{Method}
\subsection{UI-Copilot}
\paragraph{Problem Definition.} We formulate GUI automation as a sequential decision-making problem. Given a task instruction $I$ and initial screen state $S_0$, the agent generates a sequence of actions $\{a_1, a_2, \ldots, a_T\}$ to complete the task. At each step $t$, the agent observes the current screenshot state $S_t$ and samples an action $a_t$ from policy $\pi_{\theta}(a_t | I, S_t, H_{t})$, where $\theta$ denotes model parameters and $H_{t}$ represents action history. The action space $\mathcal{A}$ includes coordinate-based operations (\texttt{click}, \texttt{swipe}, \texttt{long\_press}), text-based operations (\texttt{type}, \texttt{answer}) and system-based operations (\texttt{system\_button}, \texttt{open}, \texttt{wait}, \texttt{terminate}), as shown in Table~\ref{tab:action_space}. The environment $\mathcal{E}$ transitions to
the next state according to $S_{t+1} = \mathcal{E}(S_t, a_t)$ and rollout continues until task finish or failure. 

\paragraph{Rollout Paradigm.} Given a policy agent $\mathcal{M}$ (initialized from Qwen2.5VL-7B) and a copilot model $\mathcal{M}_c$ (Qwen3-4B), we define a \textit{tool-integrated multi-turn summary rollout} paradigm for $\mathcal{M}$:
\vspace{-2mm}
\begin{Verbatim}[commandchars=\\\{\}]
\color{toolblue}<tool> \(\mathcal{T}\): tool call </tool>
\qquad\color{copilotred}\(\searrow\) call copilot model
\color{toolblue}<result> \(\mathcal{R}\): tool return </result>
\color{black}<think> \texttt{thought} </think>
\color{black}<action> \(a\) </action>
\color{purple}<summary> summary </summary>
\end{Verbatim}

\noindent where the prompt is illustrated in Figure~\ref{fig:prompt_ui_copilot}. The sampling policy is then written as
\begin{equation}
a_t, \textcolor{purple}{\texttt{summary}_t},\textcolor{toolblue}{\mathcal{T}_t},\texttt{thought}_t \sim \pi_{\theta}(\cdot | I, S_t, H_{t})
\label{eq:policy_sampling}
\end{equation}
where $\textcolor{toolblue}{\mathcal{T}_t} \in \{\texttt{Calculator}, \texttt{Retriever}, \texttt{none}\}$ denotes the role assigned to the copilot model $\mathcal{M}_c$ at step $t$. For memory-intensive tasks, the \textbf{\texttt{Retriever}} is activated using $\texttt{prompt}_R$ in Figure~\ref{fig:prompt_retriever}. It takes as input the history knowledge $\mathcal{K}$ (stored as a JSON file), the task instruction $I$, and progress summaries $\textcolor{purple}{\texttt{summary}_{<t}}$, and returns tool results to the policy agent in textual form: $\textcolor{toolblue}{\mathcal{R}_t}=\mathcal{M}_c(\texttt{prompt}_R,\mathcal{K},I,\textcolor{purple}{\texttt{summary}_{<t}})$. For numerical calculation tasks, the \textbf{\texttt{Calculator}} is invoked with $\texttt{prompt}_C$ in Figure~\ref{fig:prompt_calculator} to generate executable Python code. The generated code is then executed by a Python interpreter, and the resulting output is returned to the rollout process: $\textcolor{toolblue}{\mathcal{R}_t}=\texttt{PythonExecutor}(\mathcal{M}_c(\texttt{prompt}_C,I,\textcolor{purple}{\texttt{summary}_{<t}}))$.

\paragraph{Memory Decoupling.} Existing agents~\citep{lu2025uis1,ye2025mobileagentv3fundamentalagentsgui} maintain the full reasoning content during multi-step rollouts, updating the history as $H_t = H_{t-1} \cup \{a_{t-1}, \texttt{thought}_{t-1}\}$,
which we refer to as a \textbf{multi-turn context (MC)}.  
However, it could lead to progress confusion (e.g., redundant steps, disordered sub-task execution, or premature termination) due to content overload, as illustrated in Figure~\ref{fig:motivation} and~\ref{fig:error_type}. 
To mitigate this, we propose a \textbf{multi-turn summary (MS)} paradigm that decouples progress tracking from detailed reasoning. In the dialogue history, we maintain only a concise summary reporting current completion status (e.g., \textit{"I have finished sub-task A"}): $H_t = H_{t-1} \cup \{a_{t-1}, \texttt{summary}_{t-1}\}$.
The full reasoning content $\texttt{thought}_{t-1}$, which includes the agent's planning and explicit history observations (e.g., \textit{"The stock price is 45 dollars"}), is stored locally in a text file $\mathcal{K}$: $\mathcal{K} = \mathcal{K} \cup \texttt{thought}_{t-1}$.
This decoupling reduces context overload in the multi-turn dialogue while preserving detailed information for retrieval or numerical reasoning.
\begin{figure*}[t]
\centering
  \includegraphics[width=\textwidth]{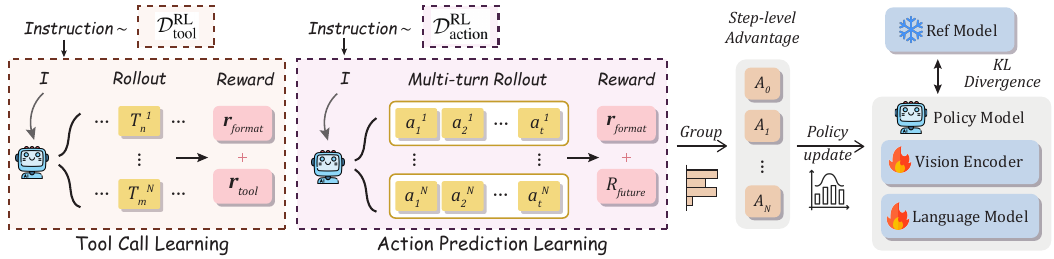}
  \caption{\textbf{Overview of \methodname Pipeline.} Policy model jointly learns tool invocations and multi-turn action prediction.}
  \vspace{-3mm}
  \label{fig:method}
\end{figure*}
\paragraph{Conclusion.} 
We formalize a \emph{multi-turn summary} interaction paradigm that leverages a copilot model as tools. This offers several advantages: \textbf{1) Lightweight Context Window.} By maintaining only concise progress summaries in the dialogue, the agent can focus on task execution, reducing the risk of planning hallucinations. \textbf{2) Decoupled Memory.} Detailed observations are stored locally and retrieved on demand, mitigating information loss and memory hallucination. \textbf{3) Efficient Inference.} Unlike previous agent workflows~\citep{wang2024mobileagent-2}, the agent selectively invokes external models, simplifying the execution pipeline.

\subsection{Dataset Curation}

\begin{figure}[h]
  \includegraphics[width=0.48\textwidth]{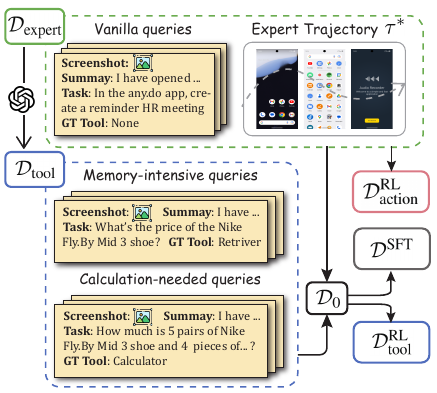}
  \caption{\textbf{Training Dataset Curation Pipeline.}}
  \vspace{-3mm}
  \label{fig:data_process}
\end{figure}

We collect $N$ diverse, human-annotated trajectories $\tau^* = \{(S_1^*, a_1^*), \dots, (S_T^*, a_T^*)\}$ from AndroidControl~\citep{li2024androidcontrol}. Further, we use GPT-4o to synthesize tool call content $\texttt{tool}_t^*$, reasoning content $\texttt{thought}_t^*$ and summary $\texttt{summary}_t^*$ for each step $t$ in each trajectory, forming the expert dataset $\mathcal{D}_\text{expert}=\{\tau^*_i\}_{i=1}^N$. For tool call learning, we use GPT-4o~\citep{hurst2024gpt} to form memory-intensive or calculation-needed queries based on $\mathcal{D}_\text{expert}$, named as $\mathcal{D}_\text{tool}$.
Then we merge them as $\mathcal{D}_\text{0}=\mathcal{D}_\text{expert}\cup\mathcal{D}_\text{tool}$, then randomly split $\mathcal{D}_\text{0}$ into $\mathcal{D}^\text{SFT}$, $\mathcal{D}_\text{tool}^\text{RL}$ and $\mathcal{D}_\text{action}^\text{RL}$ respectively for SFT and RL training. This process is detailed in Figure~\ref{fig:data_process}.

\subsection{Tool-Integrated Policy Optimization}
Our GUI agent $\mathcal{M}$ is efficiently trained on the pseudo-labeled $\mathcal{D}_0$, during which the copilot model $\mathcal{M}_c$ is not involved.
\paragraph{Cold Start.} We initialize the policy via SFT on Qwen2.5VL-7B using $\mathcal{D}^{\text{SFT}}$. Training is performed with a standard cross-entropy loss for next-token prediction, which enables both format learning and behavior cloning from expert trajectories.
\begin{equation}
\begin{aligned}
\mathcal{L}_{\mathrm{SFT}}(\theta)
&= - \mathbb{E}_{(I, \{S_t\}, \{a_t\}) \sim \mathcal{D}^{\mathrm{SFT}}} \\
&
\Bigg[
\sum_{t=1}^{T}
\log \pi_{\theta}\!\left(a_t \mid I, S_t, H_t\right)
\Bigg]
\end{aligned}
\end{equation}

\paragraph{Decoupled Sampling.} We model the agentic reasoning process in Equation~\ref{eq:agentic_rl_sampling} as explicit tool invocation, then our sampling is written as:
\vspace{-3mm}
\begin{equation}
\begin{aligned}
P_\theta(\mathcal{T}, a \mid I; \mathbb{T})& = 
\underbrace{\prod_{t=1}^{T} P_\theta(\mathcal{T}_t \mid H_{t}, I; \mathbb{T})}_{\text{Tool Calling}} 
\cdot 
\\[-2ex]
&\underbrace{\prod_{t=1}^{T} P_\theta(a_t \mid a_{<t}, H_t, I; \mathbb{T}))}_{\text{Action Generation}}
\label{eq:our_sampling}
\end{aligned}
\end{equation}
where $\mathbb{T}$ denotes the set of available tools. 

Accordingly, we adopt a decoupled strategy that separates (i) tool-calling rollouts from action-generation rollouts, and (ii) tool-call learning from action learning, depending on the source of instruction $I$. 
For $I\sim \mathcal{D}_\text{tool}^\text{RL}$, we only conduct single-turn tool prediction, conditioned on off-policy history $H_t^*=\texttt{summary}_{<t}^*$.
For $I\sim \mathcal{D}_\text{action}^\text{RL}$, we follow \citet{lu2025uis1} and conduct multi-turn action prediction, conditioned on self-generated history $H_t^\pi=\texttt{summary}_{<t}^\pi$. Equation~\ref{eq:our_sampling} is modified as:
\vspace{-1mm}
\[
\begin{aligned}
P_\theta
\approx
\begin{cases}\small
\displaystyle
\prod_{t=1}^{T}
P_\theta\!\left(\mathcal{T}_t \mid H_t^{*}, I; \mathbb{T}\right),
\text{if } I \sim \mathcal{D}_{\text{tool}}^{\text{RL}}, \\
\displaystyle
\prod_{t=1}^{T}
P_\theta\!\left(a_t \mid a_{<t}, H_t^\pi, I\right),\text{if } I \sim \mathcal{D}_{\text{action}}^{\text{RL}}
\end{cases}
\end{aligned}
\]
\vspace{-1mm}

\begin{table*}[t]

\centering
\setlength{\tabcolsep}{3pt}
\resizebox{\textwidth}{!}{%
\begin{tabular}{l>{\itshape}cICC|CC|CC|CCICC|CC|CCICC} 
\hline\thickhline
\rowcolor{gray!20} & 
& \multicolumn{8}{cI}{\textbf{\#Cross App}} 
& \multicolumn{6}{cI}{\textbf{Difficulty Level}} & &\\
\rowcolor{gray!20}
\textbf{Models}& \textbf{Type}& 
\multicolumn{2}{c|}{\textit{1 App}}
& \multicolumn{2}{c|}{\textit{2 App}}
& \multicolumn{2}{c|}{\textit{3 App}}
& \multicolumn{2}{cI}{\textit{4 App}}
& \multicolumn{2}{c|}{\textit{Easy}}
& \multicolumn{2}{c|}{\textit{Med}}
& \multicolumn{2}{cI}{\textit{Hard}}
& \multicolumn{2}{c}{\textit{Avg}} \\
\rowcolor{gray!20}

& & 
\texttt{p@1} & \texttt{p@3}
& \texttt{p@1} & \texttt{p@3}
& \texttt{p@1} & \texttt{p@3}
& \texttt{p@1} & \texttt{p@3}
& \texttt{p@1} & \texttt{p@3}
& \texttt{p@1} & \texttt{p@3}
& \texttt{p@1} & \texttt{p@3}
& \texttt{p@1} & \texttt{p@3} \\
\hline\hline

CogAgent & AO
& 0.0 & 0.0 & 0.0 & 0.0 & 0.0 & 0.0 & 0.0 &0.0 
& 0.0 & 0.0 & 0.0 & 0.0 & 0.0 & 0.0
& 0.0 & 0.0 \\

Qwen2.5VL-7B* & AT
& 14.3 & 17.9 & 1.8 & 1.8 & 0.0 & 0.0 & 0.0 & 0.0
& 10.4 & 12.5 & 0.0 & 0.0 & 0.0 & 0.0
& 3.9 & 4.7 \\

UI-Venus-7B & AT
&\underline{21.4} & 28.6 & 1.8 & 1.8 & 0.0 & 2.9 & 0.0 & 0.0 
& \underline{14.6} & 20.8 & 0.0 & 0.0 & 0.0 & 0.0
& 5.5 & 7.8 \\

UI-S1-7B* & MC
& 17.9 & 21.4 & \underline{3.6} & \underline{3.6} & 0.0 & 0.0 & 0.0 & 0.0
& 12.5 & 14.6 & \underline{2.6} & \underline{2.6} & 0.0 & 0.0
& 5.5 & 6.2 \\

UI-TARS-1.5-7B & MC
& 14.3 & 21.4 & 0.0 & 1.8 & 0.0 & 2.9 & 0.0 & 0.0
& 8.3 & 16.7 & 0.0 & 0.0 & 0.0 & 0.0
& 3.1 & 6.2 \\

GUI-Owl-7B & MC
& \underline{21.4} & \underline{35.7} & 1.8 & 1.8 & \underline{2.9} & \underline{5.9} & 0.0 & 0.0
& \underline{14.6} & \underline{22.9} & 0.0 & 2.4 & \underline{2.6} & \underline{2.6}
& \underline{6.2} & \underline{10.2} \\

\rowcolor{lightblue}\textbf{\modelname}* & TC
& \textbf{42.9} & \textbf{50.0}
& \textbf{12.5} & \textbf{16.1}
& \textbf{5.9}  & \textbf{8.8}
& 0.0  & \textbf{10.0}
& \textbf{29.2} & \textbf{33.3}
& \textbf{13.2 }& \textbf{18.4}
& \textbf{4.8}  & \textbf{7.1}
& \textbf{16.4} & \textbf{20.3} \\

\hline
\hline
Mobile-Agent-V2 & MW
& 14.3 & 17.9 & 0.0 & 0.0 & 0.0 & 0.0 & 0.0 & 0.0
& 8.3 & 10.4 & 0.0 & 0.0 & 0.0 & 0.0
& 3.1 & 3.9 \\
SeeAct & MW
& 10.7 & 25.0 & 0.0 & 0 & 0.0 & 0 & 0.0 & 0
& 6.2 & 12.5 & 0.0 & 2.4 & 0.0 & 0.0
& 2.3 & 5.5 \\
AppAgent & MW
& 14.3 & 42.9 & 0.0 & 0.0 & 0.0 & 0.0 & 0.0 & 0.0
& 8.3 & 22.9 & 0.0 & 2.4 & 0.0 & 0.0
& 3.1 & 9.4 \\
Mobile-Agent-E & MW
& 25.0 & 42.9 & 0.0 & 1.8 & 0.0 & 0.0 & 0.0 & 0.0
& 12.5 & 22.9 & 2.4 & 2.4 & 0.0 & 2.6
& 5.5 & 10.2 \\
T3A & MW
& 42.9 & 60.7 & 16.1 & 37.5 & 23.5 & 38.2 & 0.0 & 30.0
& 31.2 & 45.8 & 16.7 & 45.2 & 18.4 & 34.2
& 22.7 & 42.2 \\
M3A & MW
& 46.4 & 64.3 & 28.6 & 41.1 & 29.4 & 44.1 & 30.0 & 50.0
& 39.6 & 47.9 & 35.7 & 50.0 & 21.1 & 44.7
& 32.8 & 47.7 \\
Agent-S2 & MW
& 50.0 & 78.6 & 19.6 & 35.7 & 26.5 & 52.9 & 10.0 & 30.0
& 41.7 & 64.6 & 19.0 & 42.9 & 18.4 & 36.8
& 27.3 & 49.2 \\
\hline\thickhline
\end{tabular}%
}
\caption{\textbf{Results on MemGUI-Bench.} \texttt{p@k} denotes pass@k. Results marked with * are evaluated by ourselves with tool usage, while the remaining results are reported from the benchmark paper, where pass@3 is tested with long-term memory. The best performance in each column is highlighted in \textbf{bold}, and the second best is \underline{underlined}. \textit{\textbf{AO}} denotes \underline{\textbf{A}}ction-\underline{\textbf{O}}nly rollout without history. \textit{\textbf{AT}} denotes \underline{\textbf{A}}ction-\underline{\textbf{T}}hought history management. \textit{\textbf{MC}} denotes \underline{\textbf{M}}ulti-turn \underline{\textbf{C}}ontext. \textit{\textbf{TC}} denotes our \underline{\textbf{T}}ool-Integrated multi-turn \underline{\textbf{C}}ontext rollout. \textit{\textbf{MW}} denotes \underline{\textbf{M}}ulti-agent \underline{\textbf{W}}orkflow, with Gemini-2.5 Pro~\citep{comanici2025gemini2.5} as planning agents.}
\vspace{-3mm}
\label{tab:memgui_bench}
\end{table*}
\paragraph{Tool Call Learning.} Unlike agentic reasoning that provides tool feedback from environment, we compute rule-based reward. For $i$-th rollout,
\begin{equation}
R_t^i = 0.1 \cdot r_{\text{format}} 
        + 0.9 \cdot \mathbb{I}_{[r_{\text{format}}=1]} \cdot r_\text{tool}
\end{equation}

\paragraph{Action Prediction Learning.} For multi-turn execution learning, we train on a dataset without tool calling. First, we compute the step-wise reward as:
\begin{equation}
\begin{aligned}
r_t^i ={} & 0.1 \cdot r_{\text{format}} 
        + 0.4 \cdot \mathbb{I}_{[r_{\text{format}}=1]} \cdot r_{\text{type}} \\
        & + 0.5 \cdot \mathbb{I}_{[r_{\text{format}} \cdot r_{\text{type}}=1]} \cdot r_{\text{acc}}
\end{aligned}
\end{equation}
where all the rewards are defined in Appendix~\ref{sec:reward_definition}.
Then we introduce discounted future reward $R_t^{i} = \sum_{k=t}^{t_{\mathrm{end}}} \gamma^{,k-t} r_k^{i}$
and compute the advantage for the individual tokens using the normalized reward $R_i$: $ A_{i,t} = \frac{R_t^i - \text{mean}(\{R_t^i\}_{i=1}^G)}{\text{std}(\{R_t^i\}_{i=1}^G)}$, where $G$ is the total number of samples within a group, which is set as 8.

\noindent The training objective of \methodname is:
\vspace{-2mm}
\begin{equation}
\begin{split}
\mathcal{J}_{\text{\methodname}}(\theta) ={}
\mathbb{E}_{\substack{ I \sim \mathcal{D}^\text{RL},\{o_{i,t}\}_{1,1}^{G,T}\\ {\sim} \pi_{\theta_{\text{old}}}(\cdot\mid I)}}
\frac{1}{K}
\sum_{i=1}^{G}
\sum_{t=1}^{T}
\sum_{k=1}^{|o_{i,t}|}
\\
\min\!\left(
    \rho(\theta) A_{i,t},
    \text{clip}(\rho(\theta), 1 \pm \epsilon)  A_{i,t})
\right)
\\
 {}- \beta\, D_{KL}\!\left(\pi_\theta \,\|\, \pi_{\text{ref}}\right)
\end{split}
\end{equation}

\noindent where $\mathcal{D}^\text{RL}=\mathcal{D}_\text{tool}^\text{RL} \cup\mathcal{D}_\text{action}^\text{RL}$ denotes the RL dataset, $K$ is the total number of tokens, $\rho(\theta) = \frac{\pi_\theta(o_{i,t,k} | I, o_{i,t,<k})}{\pi_{\theta_{\text{old}}}(o_{i,t,k} | I, o_{i,t,<k})}$ is the importance sampling ratio, and $\beta$ controls the KL penalty strength. To ensure effective learning, we enforce minimum advantage variance: $\sigma(\{A_{i,t}\}) > \eta$ ($\eta$ set as 0.3), performing dynamic sampling until this threshold is met.

\section{Experiment}
\subsection{Experiment Setup}
\paragraph{Baselines.} To comprehensively assess the performance of UI-Copilot, we include three kinds of baselines: (1) advanced proprietary models, including GPT-4o~\citep{hurst2024gpt} and Claude~\citep{anthropic2024claudecu}; (2) SOTA open-source models, such as AgentCPM-GUI~\citep{zhang2025agentcpm}, GUI-Owl~\citep{ye2025mobileagentv3fundamentalagentsgui}, UI-TARS-1.5~\citep{qin2025uitars} and UI-Venus~\citep{gu2025uivenus}; (3) multi-agent workflows, such as Mobile-Agent-E~\citep{wang2025mobileagent-e}, Mobile-Agent-V2~\citep{wang2024mobileagent-2} and Agent-S2~\citep{Agent-S2}.
\begin{table*}[t]

\centering
\resizebox{\textwidth}{!}{
\begin{tabular}{lICCC|CCC||CC|CC|C}
\hline\thickhline
\rowcolor{gray!20} \textbf{Models} & \multicolumn{3}{c|}{\textbf{AC-High}} 
 & \multicolumn{3}{c||}{\textbf{GUI Odyssey}} 
 & \multicolumn{2}{c|}{\textbf{AC-Real}}
 & \textbf{Wob}
 & \textbf{AW}
 & \textbf{\textit{Avg}} \\
 
\rowcolor{gray!20} & \textit{TM} & \textit{GR} & \textit{SR} 
 & \textit{TM} & \textit{GR} & \textit{SR}
 & \textit{PG} & \textit{TSR}
 & \textit{SR}
 & \textit{SR}
 & \textbf{\textit{SR}} \\ 
\hline\hline

\multicolumn{12}{l}{\textcolor{gray!100}{\textit{Closed-source Models}}} \\

Claude-CU (SoM)~\citep{anthropic2024claudecu} 
& 63.7 & 0.0 & 12.5 & 60.9 & 0.0 & 3.1 & -- & -- & -- & 27.9 
& \textit{--} \\
GPT-4o (SoM)~\citep{hurst2024gpt} 
& 66.3 & 0.0 & 20.8 & 34.3 & 0.0 & 3.3 & -- & -- & \textbf{62.0} & \underline{34.5} 
& \textit{--} \\
\hdashline

\multicolumn{12}{l}{\textcolor{gray!100}{\textit{Open-source Models}}} \\
Qwen2VL-2B~\citep{wang2024qwen2vl} 
& 42.3 & 18.7 & 13.6 & 24.4 & 12.2 & 12.6 & 2.0 & 1.0 & 20.8 & 0.0 
& \textit{7.3} \\

ShowUI-2B~\citep{lin2024showuivisionlanguageactionmodelgui} 
& 41.8 & 32.8 & 19.7 & 34.8 & 24.6 & 21.4 & 6.8 & 2.6 & 27.1 & 7.0 
        & \textit{11.7} \\
OS-Genesis-7B~\citep{sun2024genesis}
& 65.9 & -- & 44.4 & 11.7 & -- & 3.6 & 7.6 & 3.0 & 19.8 & 17.4 
& \textit{13.4} \\
OS-Atlas-7B~\citep{wu2024atlas} 
& 57.4 & 54.9 & 29.8 & 60.4 & 39.7 & 27.0 & 14.3 & 8.6 & 35.2 & 12.1 
& \textit{18.6} \\

Qwen2.5VL-3B~\citep{bai2025qwen2_5vl} 
& 47.8 & 46.5 & 38.9 & 37.4 & 26.5 & 26.7 & 3.4 & 1.4 & 24.1 & 5.0 
& \textit{10.2} \\

Qwen2.5VL-7B~\citep{bai2025qwen2_5vl} 
& 62.2 & 72.5 & 52.7 & 67.4 & 56.3 & 52.4 & 17.4 & 9.8 & 54.0 & 22.0 
& \textit{28.6} \\

UI-R1-3B~\citep{lu2025uir1} 
& 57.9 & 55.7 & 45.4 & 52.2 & 34.5 & 32.5 & 8.4 & 4.1 & 26.1 & 8.2
& \textit{12.8} \\

UI-R1-7B~\citep{lu2025uir1} 
& 72.4 & 62.8 & 54.2 & 67.1 & 41.3 & 43.5 & 16.9 & 10.8 & 45.2 & 15.1 
& \textit{23.7} \\

AgentCPM-GUI-8B~\citep{zhang2025agentcpm} 
& 77.7 & -- & 69.2 & \underline{90.8} & -- & \underline{75.0} & 17.1 & 10.6 & 37.8 & 16.4 
& \textit{21.6} \\

UI-S1-7B~\citep{lu2025uis1}
& 79.9 & \underline{73.4} & 68.2 & 76.3 & 61.7 & 59.5 & \textbf{32.4} & \textbf{16.3} & 60.9 & 34.0 
& \textit{\underline{37.1}} \\

UI-TARS-7B~\citep{qin2025uitars} 
& \textbf{83.7} & \textbf{80.5} & \textbf{72.5} & \textbf{94.6} & \textbf{90.1} & \textbf{87.0} & 28.1 & 14.0 & 58.7 & 33.0 
& \textit{35.2} \\

\hdashline

\multicolumn{12}{l}{\textcolor{gray!100}{\textit{Ours Models}}} \\
\rowcolor{lightblue}\textbf{UI-Copilot-3B} 
& 64.3 & 54.5 & 50.3
& 52.4 & 36.8 & 35.8
& 15.6 & 6.9
&25.9 & 15.7
& \textit{16.2} \\
\rowcolor{lightblue}\textbf{\modelname} 
& \underline{82.9} & 72.2 & \underline{71.8}
& 74.5 & \underline{63.8} & 57.2
& \underline{31.5} & \underline{15.8}
& \underline{61.2} & \textbf{39.1}*
& \textit{\textbf{38.7}} \\

\hline\thickhline
\end{tabular}
}
\caption{\textbf{Results on Other GUI Benchmarks.} * shows the result with tool calling, and UI-Copilot-7B achieves 32.2\% accuracy without tool usage. Wob denotes MiniWob++ and AW denotes AndroidWorld. Average SR is computed as average of AC-Real-TSR, Wob-SR and AW-SR. The highest value is in \textbf{bold}, the second is \underline{underlined}.
}
\vspace{-3mm}
\label{tab:other_performance}

\end{table*}
\paragraph{Benchmarks.} To highlight UI-Copilot-7B’s strengths in memory- and math-intensive tasks, we first evaluate all models on the challenging MemGUI-Bench~\citep{liu2026memgui}, which consists of 70.3\% memory-intensive and 19.5\% math-intensive tasks, with an average of 36 golden steps. We then assess UI-Copilot on the widely used dynamic benchmarks AndroidWorld and MiniWob++~\citep{rawles2024androidworld} to validate its improvements in multi-turn performance.
We additionally include \texttt{AC-Real} (referred to as SOP in UI-S1~\citep{lu2025uis1}), reporting both progress (PG) and task success rate (TSR).
We also adopt the static GUI navigation benchmarks AndroidControl~\citep{li2024androidcontrol} and GUI Odyssey~\citep{lu2024guiodyssey} to evaluate comprehensive GUI understanding under high-level instructions, with action type match accuracy (TM), grounding accuracy rate (GR) and step success rate (SR) reported. 
To further evaluate the generalization ability of \modelname, we assess grounding and low-level interaction capabilities, as reported in Table~\ref{tab:st_performance}.

\subsection{Main Results}
\paragraph{Model Comparison.} As shown in Table~\ref{tab:memgui_bench}, UI-Copilot-7B achieves SOTA performance among 7B models on the challenging MemGUI-Bench. It attains a pass@1 accuracy of 16.4\% and a pass@3 accuracy of 20.3\%, substantially outperforming strong baselines like GUI-Owl-7B and UI-TARS-1.5-7B, which achieve up to 10.2\% accuracy. Moreover, UI-Copilot-7B achieves performance comparable to agentic workflows, including Mobile-Agent-E (5.5\%), AppAgent (3.1\%), and T3A (22.7\%), highlighting the efficiency of our UI-Copilot paradigm. Notably, UI-Copilot-7B successfully solves some \textit{\textbf{Hard}} tasks that require over 40 steps or 4 Apps, underscoring its potential for long-horizon, memory-intensive GUI tasks.

\paragraph{Training Effect.} Under Tool-Integrated setting, \methodname yields substantial improvements over the base model, \textit{Qwen2.5VL-7B}, demonstrating the effectiveness of our rollout and training strategy. A detailed analysis of the improvements introduced by \methodname is provided in Figure~\ref{fig:error_type}.

\paragraph{General Performance.} As shown in Table~\ref{tab:other_performance}, both our 3B and 7B models achieve substantial improvements over their base models on AC-High and GUI Odyssey, demonstrating the effectiveness of \methodname for GUI grounding and high-level understanding. Furthermore, \modelname attains advancing performance among 7B models on dynamic benchmarks such as MiniWob++ (61.2\%) and AndroidWorld (39.1\%), with results comparable to closed-source models like GPT-4o. Taken together, these results indicate that our model serves as a comprehensive GUI agent, excelling not only in long-horizon tasks but also in general scenarios.

\begin{figure}[!b]
\centering
  \vspace{-3mm}
  \includegraphics[width=0.48\textwidth]{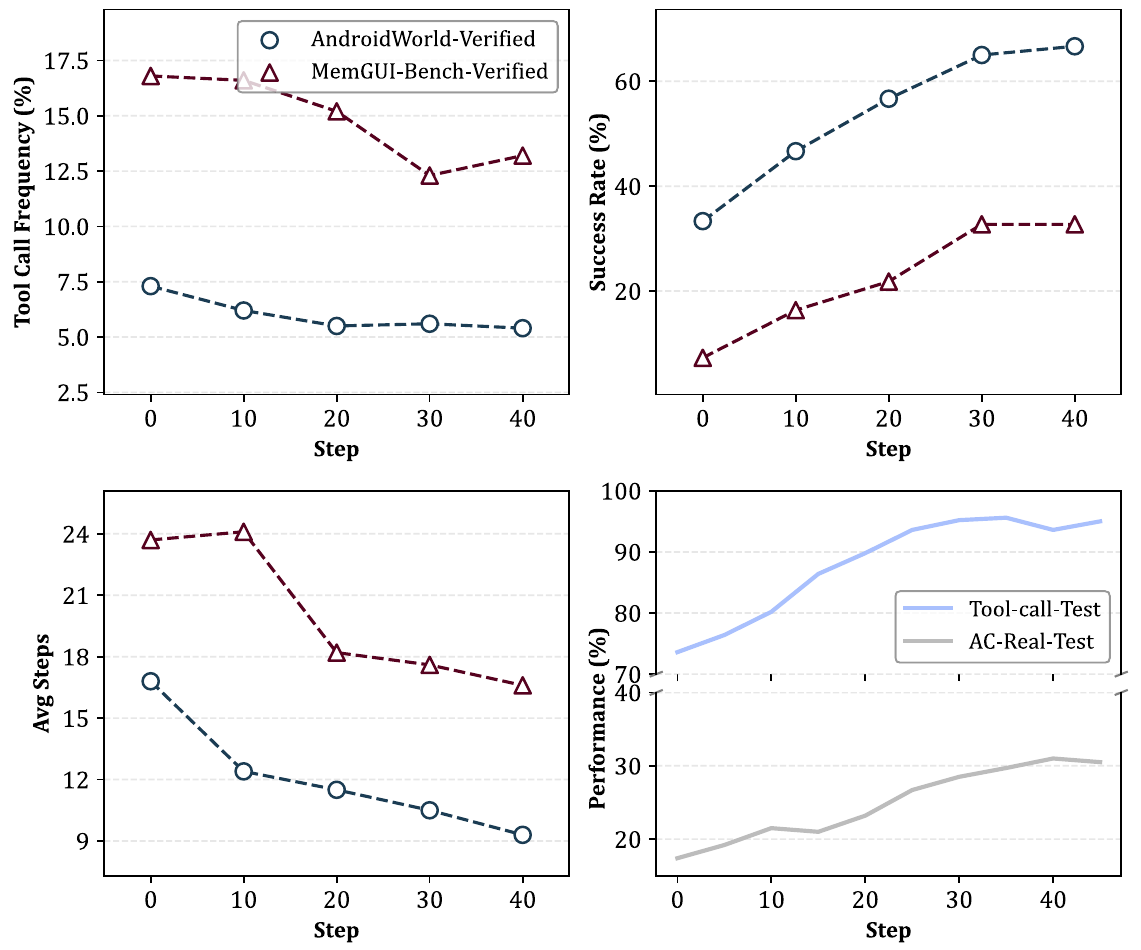}
  \caption{\textbf{Training Dynamics} of UI-Copilot-7B on our selected AndroidWorld-Verified (60 tasks), MemGUI-Bench-Verified (55 tasks) subsets, Tool-call-Test (1000 tasks from $\mathcal{D}_0$) and AC-Real-Test (1536 tasks).}
  
  \label{fig:training_dynamics}
\end{figure}
\begin{figure*}[t]
\centering

\begin{minipage}[t]{0.505\textwidth}
\centering
\resizebox{\linewidth}{!}{ 
\renewcommand{\arraystretch}{1.35} 
\setlength{\tabcolsep}{3pt} 

\begin{tabular}{>{\itshape}lccIcc|cc|cc} 
\hline\thickhline
\rowcolor{gray!20}
Method 
& \multicolumn{2}{cI}{$\mathcal{M}_c$} 
& \multicolumn{2}{c|}{MemGUI-Bench*}
& \multicolumn{2}{c|}{AndroidWorld*} & \multicolumn{2}{c}{Avg} \\

\rowcolor{gray!20}
& \texttt{Cal} & \texttt{Ret} 
& \textit{Acc(\%)$\uparrow$} & \textit{step$\downarrow$}
& \textit{Acc(\%)$\uparrow$} & \textit{step$\downarrow$}
& \textit{Acc(\%)$\uparrow$} & \textit{step$\downarrow$}\\
\hline\hline

\textit{MW} &  & \ding{51} & \heatcell{4A97C9}{\textcolor{white}{25.5}}& \heatcell{E1EDF8}{33.4}& \heatcell{08306B}{\textcolor{white}{68.3}}& \heatcell{F7FBFF}{27.8}& \heatcell{0E59A2}{\textcolor{white}{46.9}}& \heatcell{F7FBFF}{30.6} \\
\textit{MW} & \ding{51} &  & \heatcell{78B5D8}{\textcolor{white}{21.8}}& \heatcell{F7FBFF}{35.2}& \heatcell{5AA3CF}{53.3}& \heatcell{D1E2F2}{25.0}& \heatcell{62A8D2}{37.5}& \heatcell{F0F6FC}{30.1} \\
\textit{AT} &  &  & \heatcell{F7FBFF}{9.1}& \heatcell{08488F}{\textcolor{white}{20.3}}& \heatcell{F7FBFF}{35.0}& \heatcell{08306B}{\textcolor{white}{13.0}}& \heatcell{F7FBFF}{22.1}& \heatcell{083D7E}{\textcolor{white}{16.6}} \\
\textit{MC} &  &  & \heatcell{EAF2FA}{10.9}& \heatcell{083C7D}{\textcolor{white}{19.5}}& \heatcell{2D7DBB}{58.3}& \heatcell{08488F}{\textcolor{white}{14.4}}& \heatcell{8ABFDC}{34.6}& \heatcell{084286}{\textcolor{white}{16.9}} \\
\textit{MS} &  &  & \heatcell{EAF2FA}{10.9}& \heatcell{08306B}{\textcolor{white}{18.7}}& \heatcell{084991}{\textcolor{white}{65.0}}& \heatcell{08316C}{\textcolor{white}{13.1}}& \heatcell{5EA5D1}{38.0}& \heatcell{08306B}{\textcolor{white}{15.9}} \\
\textit{MS} & \ding{51} &  & \heatcell{93C4DE}{\textcolor{white}{20.0}}& \heatcell{083A7A}{\textcolor{white}{19.4}}& \heatcell{083C7D}{\textcolor{white}{66.7}}& \heatcell{083674}{\textcolor{white}{13.4}}& \heatcell{2878B8}{\textcolor{white}{43.4}}& \heatcell{083877}{\textcolor{white}{16.4}} \\
\textit{MS} &  & \ding{51} & \heatcell{78B5D8}{\textcolor{white}{21.8}}& \heatcell{08458B}{\textcolor{white}{20.1}}& \heatcell{083775}{\textcolor{white}{67.3}}& \heatcell{084185}{\textcolor{white}{14.0}}& \heatcell{1E6DB2}{\textcolor{white}{44.5}}& \heatcell{084489}{\textcolor{white}{17.1}} \\
\textit{MS} & \ding{51} & \ding{51} & \heatcell{08306B}{\textcolor{white}{36.4}}& \heatcell{083978}{\textcolor{white}{19.3}}& \heatcell{083C7D}{\textcolor{white}{66.7}}& \heatcell{083D7E}{\textcolor{white}{13.8}}& \heatcell{08306B}{\textcolor{white}{51.5}}& \heatcell{083B7B}{\textcolor{white}{16.6}} \\
\multicolumn{3}{lI}{\colorbox{gray!20}{\textit{\textbf{Copilot Model}}}}
\\ 
\multicolumn{3}{lI}{UI-Copilot-7B} & \heatcell{F7FBFF}{\textcolor{black}{23.6}} & \heatcell{CEE0F2}{\textcolor{black}{20.8}} & \heatcell{F7FBFF}{\textcolor{black}{51.7}} & \heatcell{F7FBFF}{\textcolor{black}{15.3}} & \heatcell{F7FBFF}{\textcolor{black}{37.7}} & \heatcell{F7FBFF}{\textcolor{black}{18.0}} \\
\multicolumn{3}{lI}{Qwen2.5VL-7B} & \heatcell{549FCD}{\textcolor{white}{30.9}} & \heatcell{F7FBFF}{\textcolor{black}{21.2}} & \heatcell{E2EDF8}{\textcolor{black}{53.3}} & \heatcell{3C8CC3}{\textcolor{white}{14.2}} & \heatcell{B0D2E7}{\textcolor{black}{42.1}} & \heatcell{CDE0F1}{\textcolor{black}{17.7}} \\
\multicolumn{3}{lI}{Qwen3-0.6B} & \heatcell{B9D6EA}{\textcolor{black}{27.3}} & \heatcell{3282BE}{\textcolor{white}{19.9}} & \heatcell{1C6BB0}{\textcolor{white}{63.3}} & \heatcell{125EA6}{\textcolor{white}{13.9}} & \heatcell{5CA4D0}{\textcolor{white}{45.3}} & \heatcell{1967AD}{\textcolor{white}{16.9}} \\
\multicolumn{3}{lI}{Qwen3-1.7B} & \heatcell{549FCD}{\textcolor{white}{30.9}} & \heatcell{084B93}{\textcolor{white}{19.5}} & \heatcell{3787C0}{\textcolor{white}{61.7}} & \heatcell{08306B}{\textcolor{white}{13.6}} & \heatcell{4493C7}{\textcolor{white}{46.3}} & \heatcell{08306B}{\textcolor{white}{16.6}} \\
\multicolumn{3}{lI}{Qwen3-4B} & \heatcell{08306B}{\textcolor{white}{36.4}} & \heatcell{08306B}{\textcolor{white}{19.3}} & \heatcell{08306B}{\textcolor{white}{66.7}} & \heatcell{084F99}{\textcolor{white}{13.8}} & \heatcell{08306B}{\textcolor{white}{51.6}} & \heatcell{08306B}{\textcolor{white}{16.6}} \\
\hline\thickhline
\end{tabular}
}
\caption{\textbf{Ablation Study on Inference Strategies}. * denotes the verified subset. \texttt{Cal} and \texttt{Ret} denote \texttt{Calculator} and \texttt{Retriever}, respectively. All models $\mathcal{M}$ are fine-tuned on $\mathcal{D}^{\text{RL}}$. Multi-agent workflow (MW) includes UI-Copilot-7B and Qwen3-4B. The \textit{step} includes tool invocations.}
\vspace{-3mm}
\label{tab:inference_ablation}
\end{minipage}
\hspace{0.0005\textwidth}
\begin{minipage}[t]{0.392\textwidth}
\centering
\resizebox{\linewidth}{!}{
\renewcommand{\arraystretch}{1.35} 
\setlength{\tabcolsep}{3pt} 
\begin{tabular}{ccc|cIccc} 
\hline\thickhline
\rowcolor{gray!20}
Tool & \multicolumn{2}{c|}{AC-Real} & Avg & SFT & RL & RL \\

\rowcolor{gray!20}
Acc$\uparrow$ & PG$\uparrow$ & TSR$\uparrow$ & Acc$\uparrow$ & & Tool & Action \\
\hline\hline

\heatcell{F7FBFF}{\textcolor{black}{73.6}}
& \heatcell{F0F6FD}{\textcolor{black}{17.4}}
& \heatcell{F7FBFF}{\textcolor{black}{9.88}}
& \heatcell{F7FBFF}{\textcolor{black}{33.6}}
&  &  &  \\

\heatcell{519CCC}{\textcolor{white}{86.4}}
& \heatcell{EDF4FC}{\textcolor{black}{17.6}}
& \heatcell{EEF5FC}{\textcolor{black}{10.1}}
& \heatcell{B0D2E7}{\textcolor{black}{38.0}}
& \ding{51} &  &  \\

\heatcell{083D7F}{\textcolor{white}{94.4}}
& \heatcell{F7FBFF}{\textcolor{black}{16.8}}
& \heatcell{F5F9FE}{\textcolor{black}{9.95}}
& \heatcell{6DAFD7}{\textcolor{black}{40.4}}
& \ding{51} & \ding{51} &  \\

\heatcell{1764AB}{\textcolor{white}{91.2}}
& \heatcell{206FB4}{\textcolor{white}{28.6}}
& \heatcell{3D8DC4}{\textcolor{white}{14.1}}
& \heatcell{1764AB}{\textcolor{white}{44.6}}
&  & \ding{51} & On-policy \\

\heatcell{7DB8DA}{\textcolor{black}{83.6}}
& \heatcell{08306B}{\textcolor{white}{32.4}}
& \heatcell{08306B}{\textcolor{white}{16.5}}
& \heatcell{1D6CB1}{\textcolor{white}{44.2}}
& \ding{51} &  & On-policy \\

\heatcell{539ECD}{\textcolor{white}{86.2}}
& \heatcell{9FCAE1}{\textcolor{black}{22.6}}
& \heatcell{E8F1FA}{\textcolor{black}{10.3}}
& \heatcell{82BBDB}{\textcolor{black}{39.7}}
& \ding{51} &  & Off-policy \\

\heatcell{08306B}{\textcolor{white}{95.6}}
& \heatcell{B5D4E9}{\textcolor{black}{21.5}}
& \heatcell{EBF3FB}{\textcolor{black}{10.2}}
& \heatcell{3E8EC4}{\textcolor{white}{42.4}}
& \ding{51} & \ding{51} & Off-policy \\

\heatcell{083674}{\textcolor{white}{95.0}}
& \heatcell{08478D}{\textcolor{white}{31.0}}
& \heatcell{084082}{\textcolor{white}{16.1}}
& \heatcell{08306B}{\textcolor{white}{47.4}}
& \ding{51} & \ding{51} & On-policy \\

&&&&\multicolumn{3}{c}{\colorbox{gray!20}{\textbf{$|\mathcal{D}^\text{RL}_\text{action}| : |\mathcal{D}^\text{RL}_\text{tool}|$}}}
\\

\heatcell{F7FBFF}{\textcolor{black}{91.0}} & \heatcell{4E9ACB}{\textcolor{white}{30.8}} & \heatcell{5CA4D0}{\textcolor{white}{15.8}} & \heatcell{F7FBFF}{\textcolor{black}{45.9}} & \multicolumn{3}{c}{200:600} \\
\heatcell{EEF5FC}{\textcolor{black}{91.2}} & \heatcell{08306B}{\textcolor{white}{31.5}} & \heatcell{08306B}{\textcolor{white}{16.3}} & \heatcell{C1D9ED}{\textcolor{black}{46.3}} & \multicolumn{3}{c}{300:1000} \\
\heatcell{08306B}{\textcolor{white}{95.6}} & \heatcell{F7FBFF}{\textcolor{black}{29.8}} & \heatcell{F7FBFF}{\textcolor{black}{15.2}} & \heatcell{3787C0}{\textcolor{white}{46.9}} & \multicolumn{3}{c}{600:1000} \\
\heatcell{09529D}{\textcolor{white}{95.0}} & \heatcell{2D7DBB}{\textcolor{white}{31.0}} & \heatcell{135FA7}{\textcolor{white}{16.1}} & \heatcell{08306B}{\textcolor{white}{47.4}} & \multicolumn{3}{c}{600:2000} \\
\heatcell{64A9D3}{\textcolor{white}{93.4}} & \heatcell{084F99}{\textcolor{white}{31.3}} & \heatcell{08488E}{\textcolor{white}{16.2}} & \heatcell{2575B7}{\textcolor{white}{47.0}} & \multicolumn{3}{c}{600:2400} \\
\hline\thickhline
\end{tabular}
}
\caption{\textbf{Ablations on Training Paradigms and Dataset.} Tool calling and multi-turn performance are tested on Tool-call-Test (1000 tasks) and AC-Real (1536 tasks). On/Off-policy depends on the history summary.}
\vspace{-3mm}
\label{tab:training_ablation}
\end{minipage}
\hspace{0.005\textwidth}
\begin{minipage}[t]{0.04\textwidth}
\centering
\raisebox{-0.5\height}{%
  \includegraphics[width=\linewidth]{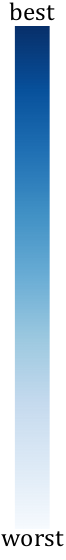}
}
\end{minipage}



\end{figure*}
\subsection{Training Dynamics}
The training dynamics in Figure~\ref{fig:training_dynamics} reveal critical insights.
\textit{\textbf{[Insight 1] Accuracy dynamics:}} Model accuracy steadily improves as training proceeds, and converges after approximately 40 training steps, indicating sufficient policy optimization.
\textit{\textbf{[Insight 2] Tool-calling dynamics:}} The frequency of tool invocations consistently decreases during training, indicating the model’s improving ability to use external tools. Notably, compared to AndroidWorld (approximately 6\% tool usage), more complex tasks such as MemGUI-Bench demand significantly higher tool utilization (nearly 13\%) and require a longer training phase for tool invocation to stabilize.
\textit{\textbf{[Insight 3] Execution efficiency dynamics:}} As training progresses, the average execution steps decrease, suggesting that RL effectively reduces redundant actions and mitigates progress confusion, leading to more efficient completion.

\subsection{Ablation Analysis}

\paragraph{Ablations on Rollout Paradigm.} As shown in Figure~\ref{tab:inference_ablation} (upper part), under the setting without tool calling, ours rollout paradigm, \textbf{Multi-turn Summary (MS)}, consistently achieves higher accuracy with fewer execution steps than Action–Thought (AT) and Multi-turn Context (MC). This indicates that MS effectively mitigates redundant actions and mitigates progress confusion during multi-turn execution. Based on the MS paradigm, we further conduct tool-set ablations. The results show that both the \texttt{Calculator} and \texttt{Retriever} contribute to improved performance on MemGUI-Bench. Notably, the full collaborative tool set achieves the best overall performance, with 51.5\% average accuracy and 16.6 average steps. It also attains competitive performance against the multi-agent workflow which invokes the copilot model at every step, demonstrating the efficiency of UI-Copilot and \methodname.

\paragraph{Ablations on Copilot Model.} We compare different copilot models in Figure~\ref{tab:inference_ablation} (bottom part). Among all models, Qwen3-4B achieves the best performance, outperforming Qwen3-0.6B, Qwen-1.7B, and MLLMs such as Qwen2.5VL-7B. This result highlights the strong capability of Qwen3-4B in context understanding and summarization, which is crucial for effective copilot assistance.
\begin{figure*}[t]
\centering
  \includegraphics[width=1\textwidth]{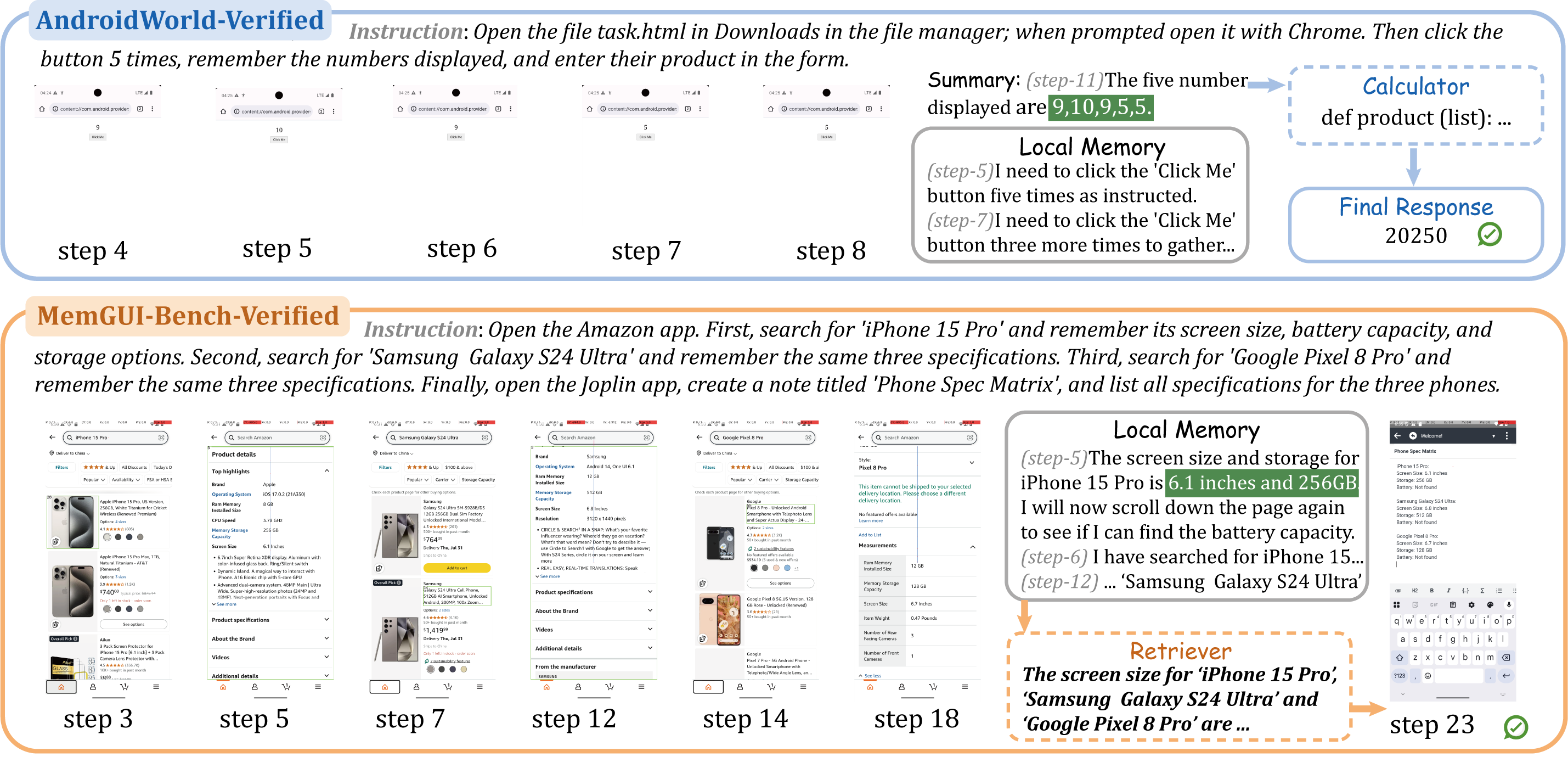}
  \caption{\textbf{Case Study.} UI-Copilot-7B successfully completes a math-related task from \sethlcolor{androidworld}\hl{AndroidWorld} (top) and a memory-related task from \sethlcolor{memguibench}\hl{MemGUI-Bench} (bottom).}
  \vspace{-3mm}

  \label{fig:case_study}
\end{figure*}
\paragraph{Ablations on TIPO.} We conduct ablations on training paradigms (TIPO) in Figure~\ref{tab:training_ablation} (upper part). The results indicate that SFT plays a critical role as a cold start, providing a reliable initialization for subsequent RL training. Furthermore, both tool call RL and action prediction RL are essential for effective tool calling (see Tool-call-Test) and stable multi-turn execution (see AC-Real), respectively. For action prediction learning, on-policy (multi-turn self-generated) histories consistently outperform off-policy (expert-collected) histories, due to the better alignment with multi-turn evaluation.
\paragraph{Ablations on Training Dataset.} 
Figure~\ref{tab:training_ablation} (bottom part) demonstrates 600:2000 as an optimal data ratio for (\textbf{$|\mathcal{D}^\text{RL}_\text{action}| : |\mathcal{D}^\text{RL}_\text{tool}|$}), which achieves a favorable balance between tool-call learning and action generation learning. Increasing the size of $\mathcal{D}^\text{RL}_\text{action}$ or $\mathcal{D}^\text{RL}_\text{tool}$ does not yield further improvements, indicating diminishing returns from additional action-level or tool-level learning samples.

\begin{figure}[!b]
\centering
  \vspace{-3mm}\includegraphics[width=0.48\textwidth]{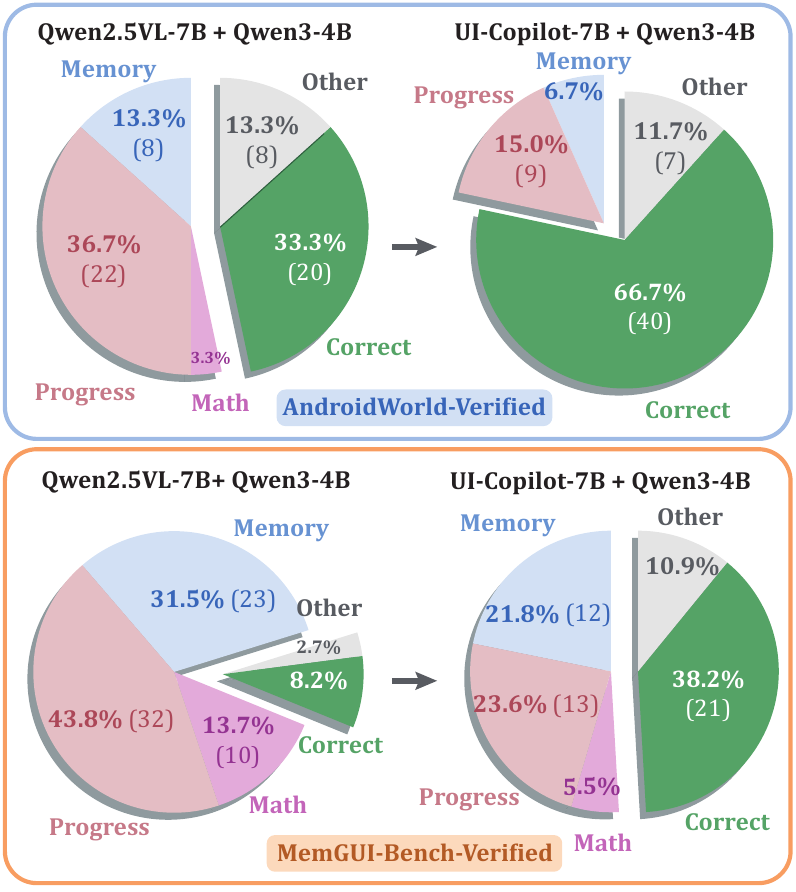}
  \caption{\textbf{Error Type Analysis with Tool Usage.} Errors are categorized into \sethlcolor{memory}\hl{Memory} Degradation, \sethlcolor{progress}\hl{Progress} Confusion, \sethlcolor{math}\hl{Math} Hallucination and \sethlcolor{other}\hl{Other} Fault.}
  \vspace{-1mm}
  \label{fig:error_type}
\end{figure}
\subsection{Case Study and Analysis}
\paragraph{Case Study.} Figure~\ref{fig:case_study} presents two challenging cases from AndroidWorld and MemGUI-Bench, which require numerical computation and information retrieval, respectively. In both scenarios, \modelname successfully invokes the Copilot model as the appropriate external tool. Specifically, in the AndroidWorld example, the Calculator generates executable code at step 9 and returns the computed results to \modelname. In the MemGUI-Bench case, the Retriever gathers critical information at steps 5, 12, and 18, enabling \modelname to correctly complete the form-filling task at step 23. Additional successful and failed cases are provided in Appendix~\ref{sec:more_cases}, further demonstrating the effectiveness of our tool-invocation and training method.

\paragraph{Error Type Analysis.} Figure~\ref{fig:error_type} illustrates several representative error types categorized using GPT-4o between \methodname (\textbf{MS with tool}) and Qwen2.5VL-7B (\textbf{AT with tool}). Among all error categories, \textit{Progress Confusion} emerges as the dominant failure mode on both benchmarks. For the more challenging MemGUI-Bench (only 8.2\% success rate orignally), MLLMs further suffer from more reasoning limitations, including \textit{Memory Degradation} and \textit{Math Hallucination}. As results demonstrate, our \modelname achieves substantial improvements over the base Qwen2.5VL-7B (+33.4\% on AndroidWorld and +30.0\% on MemGUI-Bench). From the error-type perspective, \textit{Progress Confusion} is almost halved, while \textit{Memory Degradation} and \textit{Math Hallucination} are significantly mitigated, demonstrating the effectiveness of our \methodname. Error analysis of tool invocation is shown in Figure~\ref{fig:error_type_tool}.
\section{Conclusion}
We incorporate Copilot Model as external tools and propose \methodname for tool-integrated learning, aiming to solve complex and long-horizon GUI tasks. Our \modelname achieves consistently strong performance across several challenging GUI benchmarks.
\section*{Limitations}
Currently, our tool set is limited to Calculator and Retriever. However, real-world GUI scenarios often require a broader spectrum of tools, such as web search and visual cropping. Extending our framework to support more diverse tool integrations remains an important direction for future work.




\bibliography{custom}
\clearpage
\appendix

\section{Action Space}
\label{sec:appendix}
\begin{table}[h]
\centering
\resizebox{0.48\textwidth}{!}{
\begin{tabular}{l|p{0.7\columnwidth}}
\toprule
\textbf{Action Type} & \textbf{Description} \\
\midrule
\texttt{click} & Tap a specific coordinate $(x, y)$ on the screen. \\
\texttt{long\_press} & Press and hold at $(x, y)$ for a specified duration. \\
\texttt{swipe} & Perform a swipe gesture from $(x_1, y_1)$ to $(x_2, y_2)$. \\
\texttt{answer} & Output a textual answer to the task. \\
\texttt{type} & Enter text into the currently focused input field. \\
\texttt{system\_button} & Trigger a system-level button (e.g., Home, Back). \\
\texttt{open} & Launch an APP on the device. \\
\texttt{wait} & Pause execution for a given number of seconds to allow UI changes. \\
\texttt{terminate} & Stop execution and report task success or failure. \\
\bottomrule
\end{tabular}
}
\caption{Action space in AndroidWorld automation.}
\label{tab:action_space}
\end{table}
\begin{figure}[h]
\centering
  \includegraphics[width=0.48\textwidth]{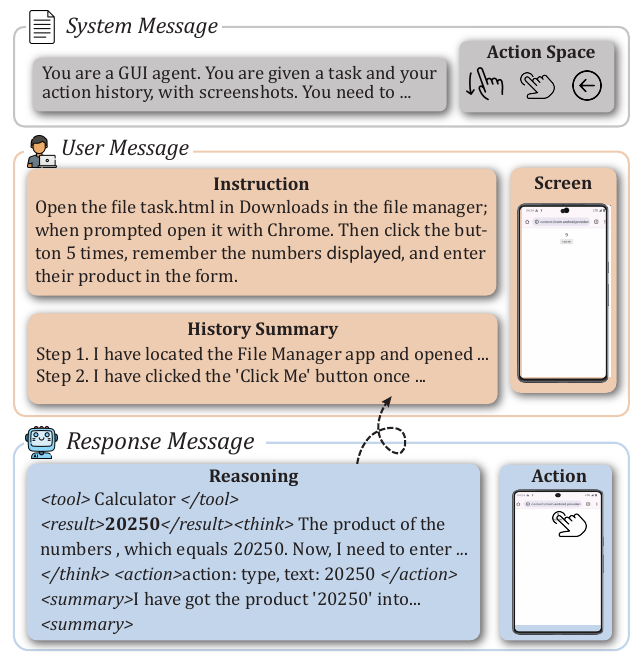}
  \caption{\textbf{Interaction example for UI-Copilot-7B.}}
  \label{fig:interaction_flow}
\end{figure}
\newpage
\section{Reward Definition}
\label{sec:reward_definition}

\subsection{Type Reward ($r_{\text{type}}$)}

$r_{\text{type}}\in\{0,1\}$ indicates whether the predicted action type matches the ground-truth action type.
Let $a^{\text{pred}}$ and $a^{\text{gt}}$ denote the predicted and ground-truth action types, respectively.
The type reward is defined as
$r_{\text{type}}=\mathbb{I}[a^{\text{pred}}=a^{\text{gt}}]$.

\subsection{Accuracy Reward ($r_{\text{acc}}$)} 

$r_{\text{acc}}\in\{0,1\}$ evaluates whether the predicted action is accurate given the ground-truth action,
conditioned on the action type being correct and after coordinate normalization.
Let $\mathbf{p}^{\text{pred}}$ and $\mathbf{p}^{\text{gt}}$ denote the predicted and ground-truth coordinates, respectively.

\begin{itemize}

\item \textbf{Wait / Terminate} 

The prediction is accurate if the action type exactly matches:
$r_{\text{acc}}=\mathbb{I}[a^{\text{pred}}=a^{\text{gt}}]$.

\item \textbf{System Button} 

Let $\text{button}^{\text{pred}}$ and $\text{button}^{\text{gt}}$ denote the predicted and ground-truth system button names.
The prediction is accurate if the button names match in a case-insensitive manner:
$r_{\text{acc}}=\mathbb{I}[\text{button}^{\text{pred}}=\text{button}^{\text{gt}}]$.

\item \textbf{Type / Answer / Key / Open} 

Let $\text{text}^{\text{pred}}$ and $\text{text}^{\text{gt}}$ denote the predicted and ground-truth input strings.
The prediction is accurate if the texts match under relaxed string matching:
$r_{\text{acc}}=\mathbb{I}[\text{text}^{\text{pred}}\sim\text{text}^{\text{gt}}]$.

\item \textbf{Swipe} 

Let $\mathbf{p}_1^{\text{pred}}, \mathbf{p}_2^{\text{pred}}$ denote the start and end points of the predicted swipe,
and $\text{dir}(\cdot,\cdot)$ be the function that infers swipe direction.
The prediction is accurate if the inferred swipe direction matches the ground truth:
$r_{\text{acc}}=\mathbb{I}[\text{dir}(\mathbf{p}_1^{\text{pred}},\mathbf{p}_2^{\text{pred}})=\text{dir}^{\text{gt}}]$.

\item \textbf{Click / Long Press} 

Let $\mathcal{B}^{\text{gt}}$ denote the enlarged ground-truth bounding box
and $\epsilon$ be a fixed distance threshold.
The prediction is accurate if the predicted point falls inside the bounding box or is sufficiently close to the ground-truth point:
$r_{\text{acc}}=\mathbb{I}[\mathbf{p}^{\text{pred}}\in\mathcal{B}^{\text{gt}}\ \lor\ \lVert\mathbf{p}^{\text{pred}}-\mathbf{p}^{\text{gt}}\rVert_2\le\epsilon]$.

\end{itemize}
\newpage
\section{Theoretical Analysis}
\label{sec:full_proof}
\subsection{Preliminaries: Agentic RL}
Agentic RL incorporates tool-call feedback during the reasoning process~\citep{tora,torl,wu2025agentic,dong2025arpo}. The rollout sampling can be decomposed as:
\vspace{-4mm}
\begin{equation}
\begin{aligned}
P_\theta(\mathcal{R}, a \mid I; \mathbb{T}) &= 
\underbrace{\prod_{t=1}^{t_{\mathcal{R}}} P_\theta(\mathcal{R}_t \mid \mathcal{R}_{<t}, I; \mathbb{T})}_{\text{Agentic Reasoning}} 
\cdot 
\\[-2ex]& \underbrace{\prod_{t=1}^{t_a} P_\theta(a_t \mid a_{<t}, \mathcal{R}, I; \mathbb{T}))}_{\text{Action Generation}},
\label{eq:agentic_rl_sampling}
\end{aligned}
\end{equation}
where $\mathbb{T}$ denotes the set of available tools, $\mathcal{R}$ is the reasoning trajectory of length $t_{\mathcal{R}}$, interleaved with tool-call feedback, and $a$ is the performed action with length $t_a$.
\subsection{Multi-turn Action Prediction Learning}
We prove in this section that why we use on-policy multi-turn rollout for action prediction learning.
\paragraph{Setup.}
Let $\trainerpi(\cdot \mid \theta)$ denote the training policy and
$\rolloutpi(\cdot \mid \theta)$ the rollout (inference) policy.  
Given a high-level instruction $I$, the deployment objective is
{\small
\[
\mathcal{J}(\theta)
=
\mathbb{E}_{I \sim p_{\mathcal{I}}} \Big[
\mathbb{E}_{(a_{1:T}, \mathcal{T}_{1:T}) \sim \rolloutpi(\cdot\mid I)}
\big[R(I, a_{1:T}, \mathcal{T}_{1:T})\big]
\Big].
\]
}

\paragraph{Single-turn Training Mismatch.}
In single-turn (ST) training, each step conditions on off-policy histories $H_t^*$, yielding
\[
\begin{aligned}
&\widehat{\nabla_\theta \mathcal{J}}_{\mathrm{ST}}
=
\mathbb{E}_{I \sim p_{\mathcal{I}}, a_t \sim \trainerpi(\cdot \mid I, S_t, H_t^*)}\\
&\qquad \Big[
\nabla_\theta \log \trainerpi(a_t \mid I, S_t, H_t^*)\, 
R(I, a_{1:T}, \mathcal{T}_{1:T})
\Big].
\end{aligned}
\]
Evaluation, however, uses self-generated histories $H_t^\pi$:
\(
a_t^{\pi} \sim \rolloutpi(\cdot \mid I, S_t, H_t^{\mathrm{MT}}),
\)
so that
\[
\widehat{\nabla_\theta \mathcal{J}}_{\mathrm{ST}}
\neq
\nabla_\theta \mathbb{E}_{(a_{1:T}^{\pi}, \mathcal{T}_{1:T}^{\pi}) \sim \rolloutpi}[R(I, a_{1:T}^{\pi}, \mathcal{T}_{1:T}^{\pi})].
\]
Equivalently,
\[
\begin{aligned}
\argmax_\theta \mathbb{E}_{a_{1:T} \sim \trainerpi}[R(I, a_{1:T})]
\;\neq\; \\
\argmax_\theta \mathbb{E}_{a_{1:T}^{\pi} \sim \rolloutpi}[R(I, a_{1:T}^{\pi})]
\end{aligned}
\]
which illustrates the biased gradient and deployment gap.

\paragraph{Multi-turn Training Alignment.}
Multi-turn (MT) training conditions on self-generated histories $H_t^\pi$ at each step, producing full trajectories
\(
(a_{1:T}^{\pi}, \mathcal{T}_{1:T}^{\pi}) \sim \rolloutpi(\cdot \mid I)
\)
and the gradient estimator
\[
\begin{aligned}
\widehat{\nabla_\theta \mathcal{J}}_{\mathrm{MT}}
&=
\mathbb{E}_{I \sim p_{\mathcal{I}}} \Big[
\sum_{t=1}^{T} \nabla_\theta \log \rolloutpi(a_t^{\pi} \mid I, S_t, H_t^\pi)\, \\
&\quad R(I, a_{1:T}^{\pi}, \mathcal{T}_{1:T}^{\pi})
\Big].
\end{aligned}
\]
By aligning training histories with rollout histories, MT training better approximates the deployment objective:
{\small
\[
\argmax_\theta \mathbb{E}_{(a_{1:T}^{\pi}, \mathcal{T}_{1:T}^{\pi}) \sim \rolloutpi}[R(I, a_{1:T}^{\pi})]
\approx
\argmax_\theta \mathcal{J}(\theta)
\]
}
reducing the train–inference mismatch.

\paragraph{Conclusion.}
Single-turn training suffers from biased gradients due to off-policy histories $H_t^*$, whereas multi-turn training uses self-generated histories $H_t^\pi$, leading to more consistent and stable optimization toward deployment-time performance.
\subsection{Decoupled Sampling in RL Training}
We consider the policy gradient
\[
\begin{aligned}
\nabla_\theta \mathcal{J}(\theta)
&= \mathbb{E}\Big[
A(I,\mathcal{T},a) \,\big(
\nabla_\theta \log P_\theta(\mathcal{T}\mid I) \\
&\qquad\; + \nabla_\theta \log P_\theta(a\mid \mathcal{T}, I)
\big)
\Big].
\end{aligned}
\]
where $A(\cdot)$ denotes the advantage function.

\paragraph{Tool-learning instructions.}
For $I\sim\mathcal{D}_{\mathrm{tool}}^{\mathrm{RL}}$, tool calls are supervised and
consistent across trajectories.
Conditioned on a fixed tool $\mathcal{T}$, the action distribution
$P_\theta(a\mid \mathcal{T}, I)$ becomes highly concentrated, yielding a small
action-level advantage:
\[
\mathbb{E}_{a\sim P_\theta(\cdot\mid \mathcal{T}, I)}
\big[
|A(I,\mathcal{T},a)|
\big]
\approx 0 .
\]
As a result, the action-related gradient term contributes negligibly and can be
ignored when estimating the policy gradient.

\paragraph{Progress-learning instructions.}
For $I\sim\mathcal{D}_{\mathrm{action}}^{\mathrm{RL}}$, no tool is required and
we explicitly fix the tool set to $\mathcal{T}=\texttt{None}$ in the prompt.
In this case, the policy reduces to action generation only, and the gradient
simplifies to
\[
\nabla_\theta \mathcal{J}_{\mathrm{prog}}(\theta)
=
\mathbb{E}\!\left[
A(I,\texttt{None},a)\,
\nabla_\theta \log P_\theta(a\mid I)
\right]
\]
where rewards and advantages are computed solely based on task progress.
\begin{figure*}[!h]
    \centering
    \begin{minipage}[t]{0.32\textwidth}
        \centering
        \includegraphics[width=\linewidth]{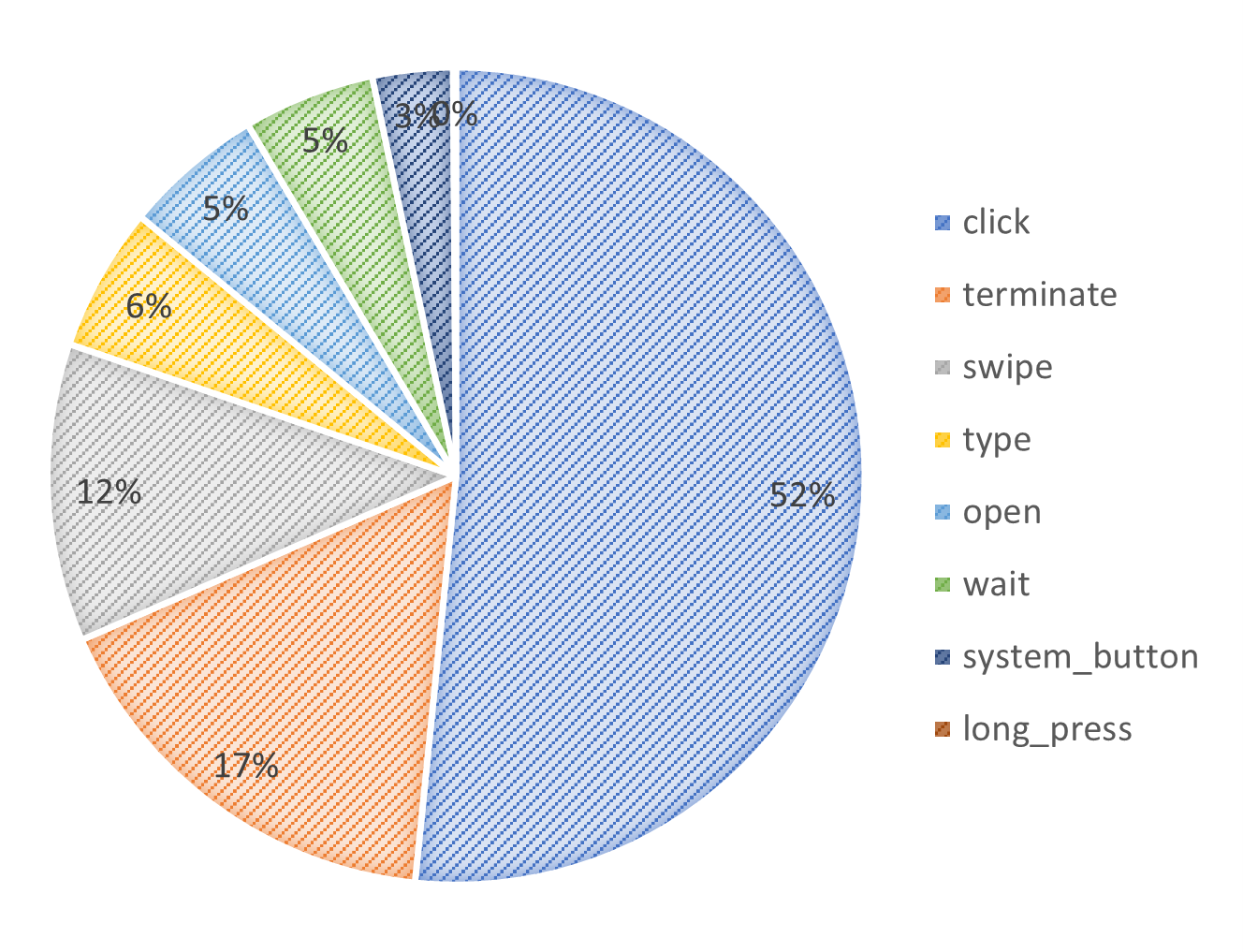}
        \caption{Action Type Distribution of $\mathcal{D}_\text{action}^\text{RL}$.}
        \label{fig:training_action_type}
    \end{minipage}
    \hfill
    \begin{minipage}[t]{0.32\textwidth}
        \centering
        \includegraphics[width=\linewidth]{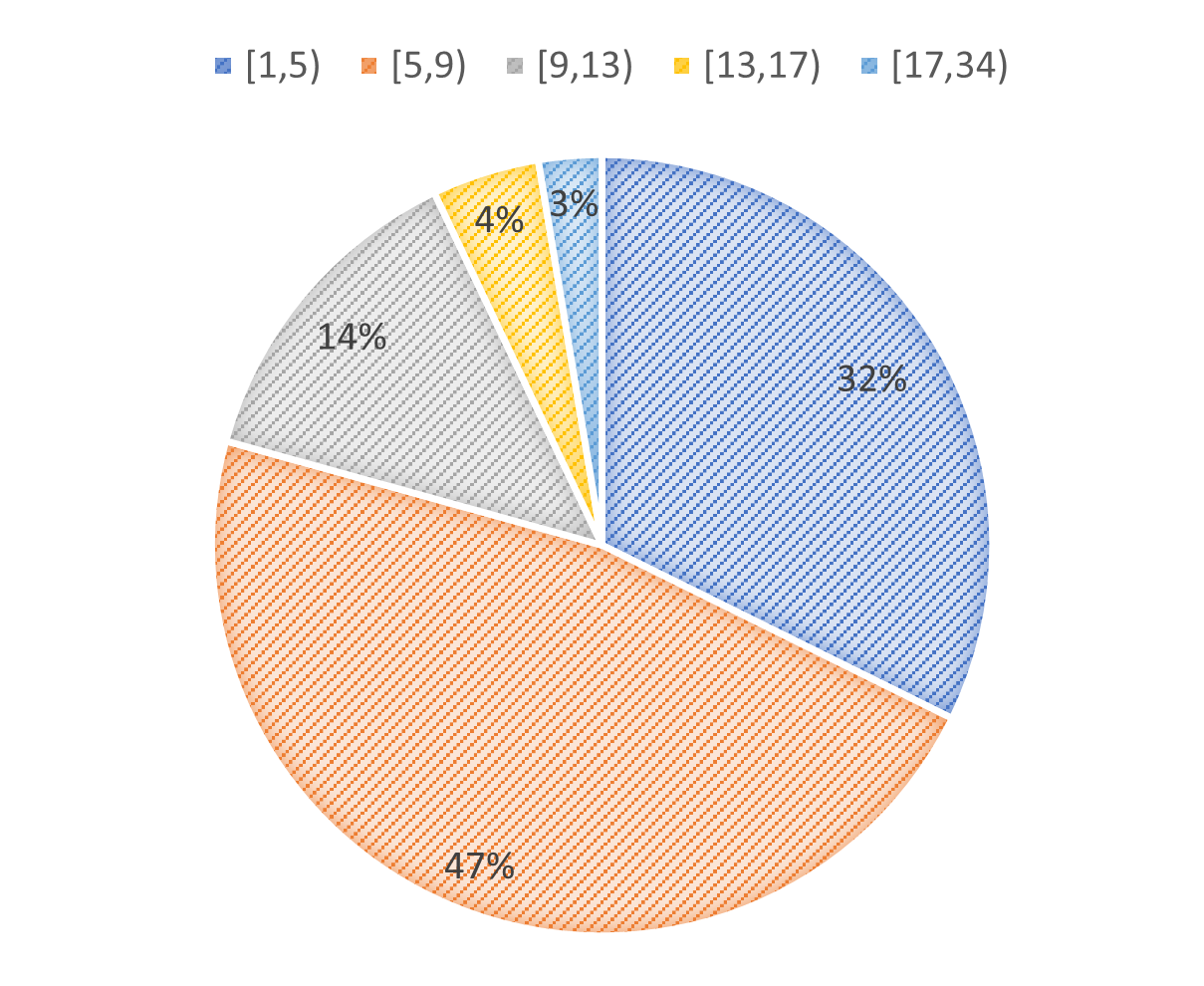}
        \caption{Trajectory Length Distribution of $\mathcal{D}_\text{action}^\text{RL}$.}
    \label{fig:training_traj_length}
    \end{minipage}
    \hfill
    \begin{minipage}[t]{0.32\textwidth}
        \centering
        \includegraphics[width=\linewidth]{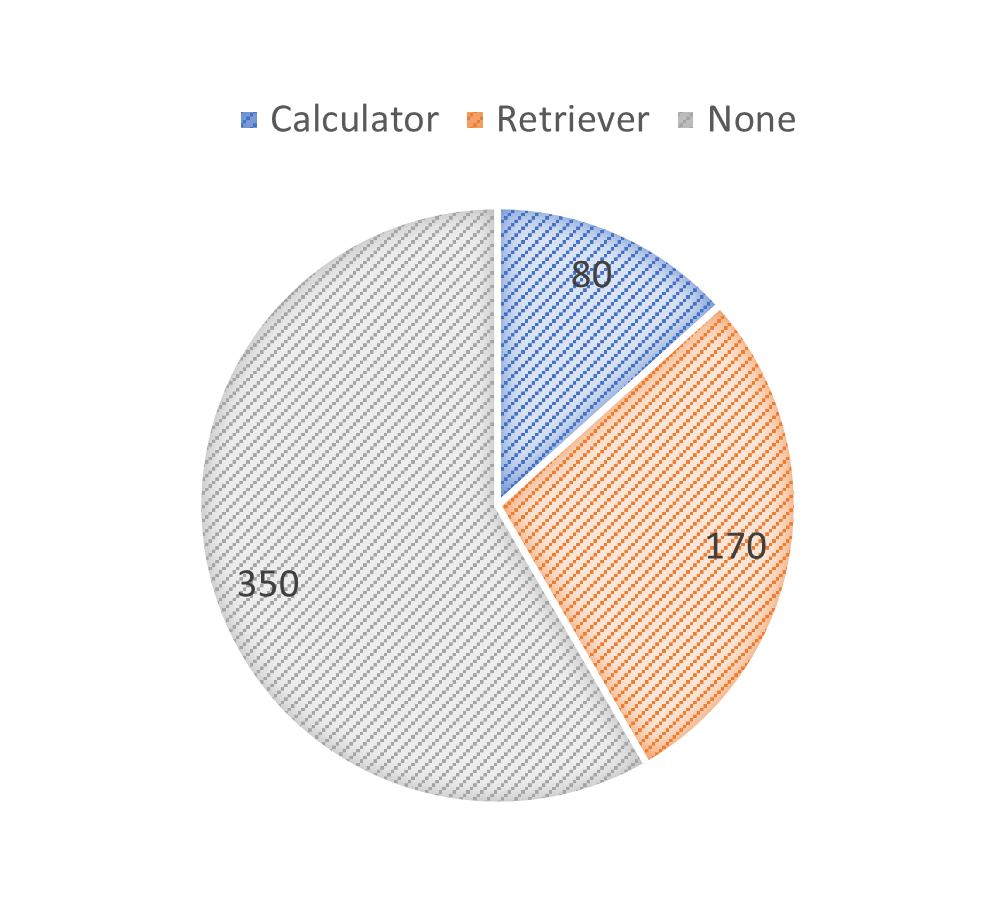}
        \caption{Tool Type Distribution of $\mathcal{D}_\text{tool}^\text{RL}$.}
    \label{fig:training_tool_type}
    \end{minipage}
\end{figure*}

\newpage

\section{Data Description}
\subsection{Training Dataset}
\paragraph{$\mathcal{D}_\text{action}^\text{RL}$ Composition.} Figures~\ref{fig:training_action_type} and~\ref{fig:training_traj_length} illustrate the distributions of action types and trajectory lengths across 2000 training trajectories in $\mathcal{D}_\text{action}^\text{RL}$, respectively. Among action types, \texttt{CLICK} is the most prevalent, followed by \texttt{TERMINATE}, which consistently serves as the final action in all successfully completed trajectories, and \texttt{SWIPE}. In terms of trajectory length, most trajectories comprise between 5 and 9 interaction steps. We provide some cases below,
\begin{Verbatim}[breaklines, breakanywhere]
Open the Zoho Meet app, view the scheduled meetings.
Go to the mobile category after closing the pop up and browse the products.
Check out all of the suggested products and compare pricing because I'm looking for a budget friendly sofa.
I want to view the recipe for Welsh Cakes in the kitchen stories app.
Open the Yahoo Mail App, Select the Starva Mail and Unmark an Email as read.
If I lose connection to the internet, I want to make the agents.txt file accessible offline in Google Drive so that I can readily access it.
I want to share "Oscar and the wolf -somebody wants" music to my friend karin.iversen@example.com via gmail.
In the HealthifyMe app, view your today activity
Add paintings to cart on the Rtistiq app.
Set 10 minutes before notification for the birthday event in gmail calendar
Im going to Deutsches Museum from Ulm city, so get the traffic update on the route of the Deutsches Museum from my location.
I want to search for some interesting activities in Hawaii.
Open the TickTick app and mark microsoft training update classes as complete.
I want to read the reviews of the Nike Fly.By Mid 3 shoe in the Nike app.
Search for Kayak mail in gmail app.
Use the gmail address karin.iversen@example.com to send Karin information about the bus that leaves at 1:55 a.m.
In the Simple Habit app, In order to improve my meditation, I would like to listen to the sound of ocean.
\end{Verbatim}
\vspace{-4mm}
\begin{figure*}[!h]
    \centering
        \begin{minipage}[t]{0.3\textwidth}
        \centering
        \includegraphics[width=\linewidth]{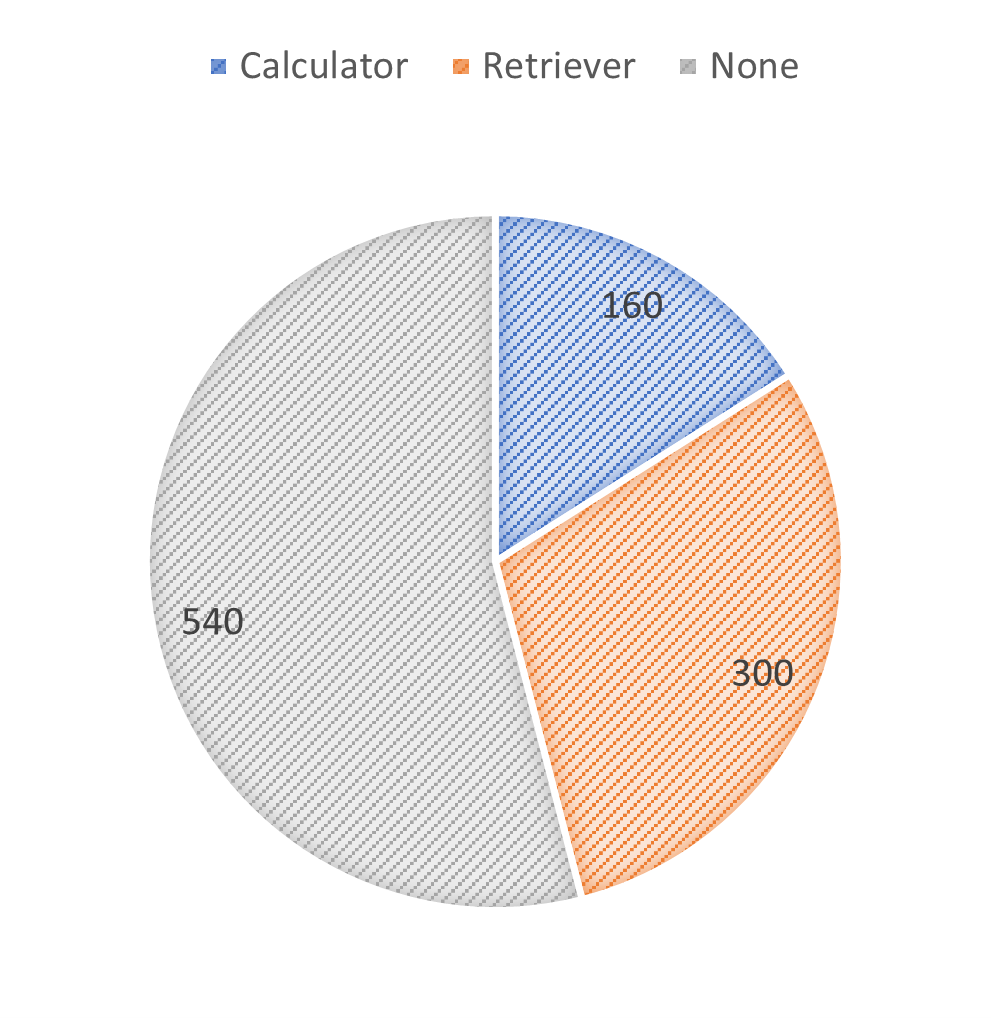}
        \caption{Tool Type Distribution of Tool-call-Test.}
    \label{fig:test_tool_type}
    \end{minipage}
    \hfill
    \begin{minipage}[t]{0.3\textwidth}
        \centering
        \includegraphics[width=\linewidth]{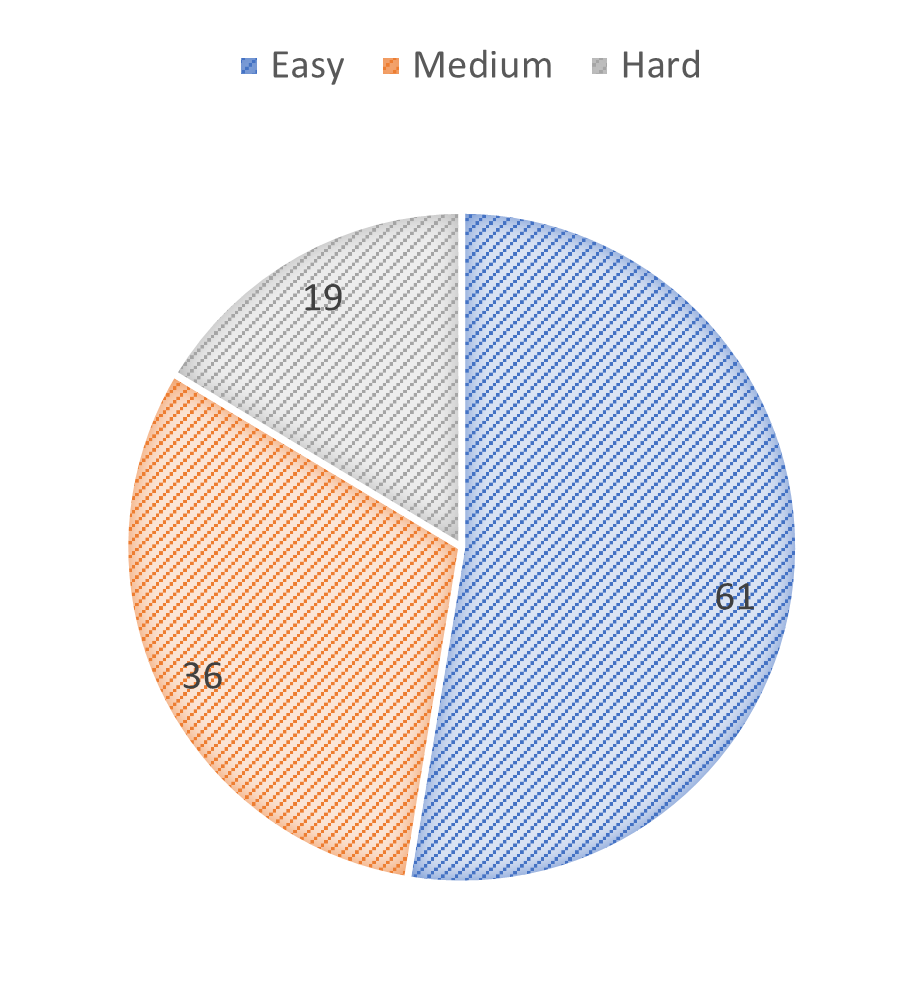}
        \caption{Difficulty Level Distribution of AndroidWorld.}
        \label{fig:androidworld_difficulty}
    \end{minipage}
    \hfill
    \begin{minipage}[t]{0.3\textwidth}
        \centering
        \includegraphics[width=\linewidth]{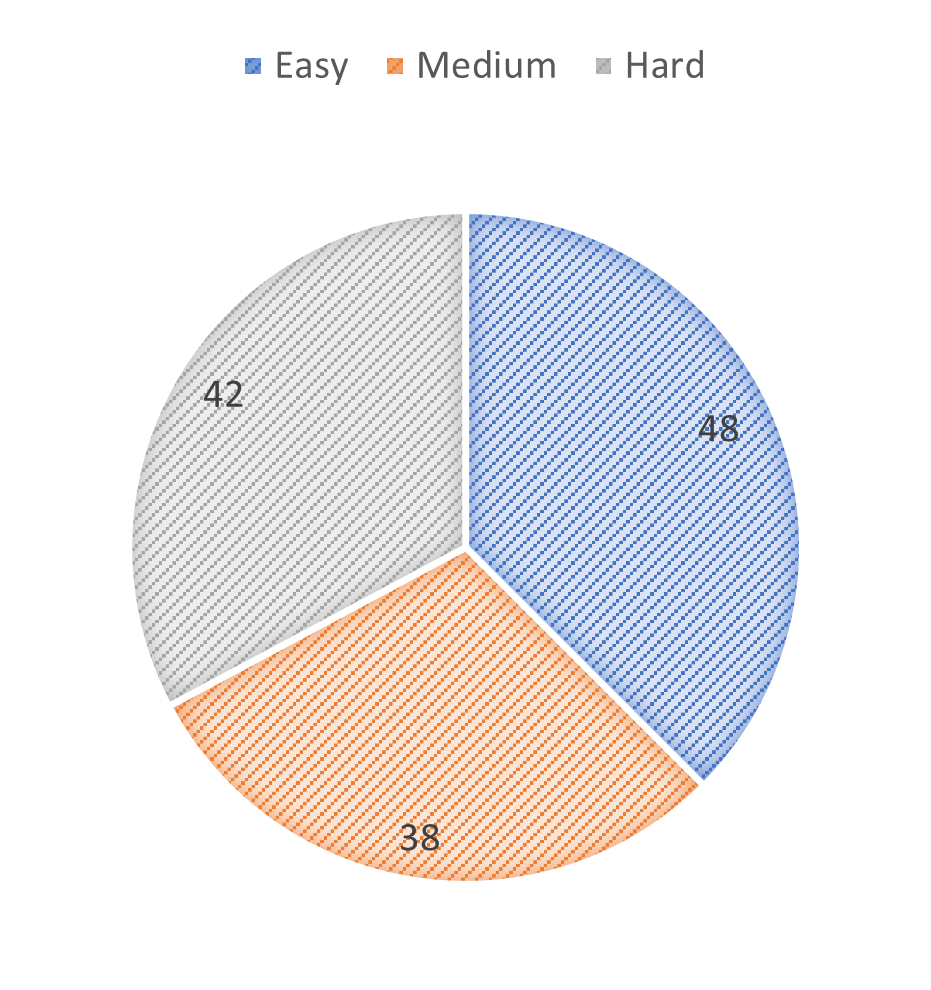}
        \caption{Difficulty Level Distribution of MemGUI-Bench.}
    \label{fig:memguibench_difficulty}
    \end{minipage}
\end{figure*}
\paragraph{$\mathcal{D}_\text{tool}^\text{RL}$ Composition.} The dataset $\mathcal{D}_\text{tool}^\text{RL}$ is constructed using GPT-4o and includes 170 memory-intensive queries, 80 calculation-required queries, and 350 tasks requiring no tool usage, as illustrated in Figure~\ref{fig:training_tool_type}. We also provide some cases below,

\vspace{-4mm}
\begin{Verbatim}[breaklines, breakanywhere]
How much would it cost to buy two pairs of Nike Fly.By Mid 3 shoes, and which option is more budget-friendly compared to similar models?
Among the suggested sofas, which one is the most budget-friendly based on their prices and features?
What is the total price after adding the Rockrider City Cycle Btwin bicycle to my cart, including any applicable discounts?
What is the total cost of all selected paintings added to the cart in the Rtistiq app?
What is the cheapest available flight from Knoxville to Hawaii, and how do prices vary across different dates?
What is the estimated travel time and cost when traveling from Los Angeles to Oakland by train?
What are the available bus options from Amsterdam Centraal to Rotterdam Centraal on October 26, and how long does each trip take?
How long will it take to travel from Ulm to the Deutsches Museum based on current traffic conditions?
How much earlier will I be notified if I set the calendar reminder to 10 minutes before the birthday event?
Which file was made available offline in Google Drive when preparing for loss of internet connection?
Which music track was shared via Gmail, and to which recipient was it sent?
Which bus departure information was sent to karin.iversen@example.com?
What route and traffic conditions were shown for traveling from Ulm to the Deutsches Museum?
Which activities were shown after searching for things to do in Hawaii?
Which Microsoft training classes were marked as complete in the TickTick app?
What account or session status was shown after signing out of the Babbel app?
\end{Verbatim}
\paragraph{Training Details.} We train Qwen2.5VL-7B with \methodname for 50 steps on 8 A100 GPUs. Each batch samples 16 prompts, with 8 rollouts per prompt. We define the maximum response length as 12288 tokens and learning rate as $1\times 10^{-6}$.
\subsection{Evaluation Dataset}
\label{sec:evaluation_data}
\paragraph{Tool-call-Test.} As shown in Figure~\ref{fig:test_tool_type}, the Tool-call-Test subset consists of 1000 tasks generated using GPT-4o and is carefully aligned with $\mathcal{D}_\text{tool}^\text{RL}$ in terms of task-type distribution.

\begin{figure}[h]
\centering
  \includegraphics[width=0.4\textwidth]{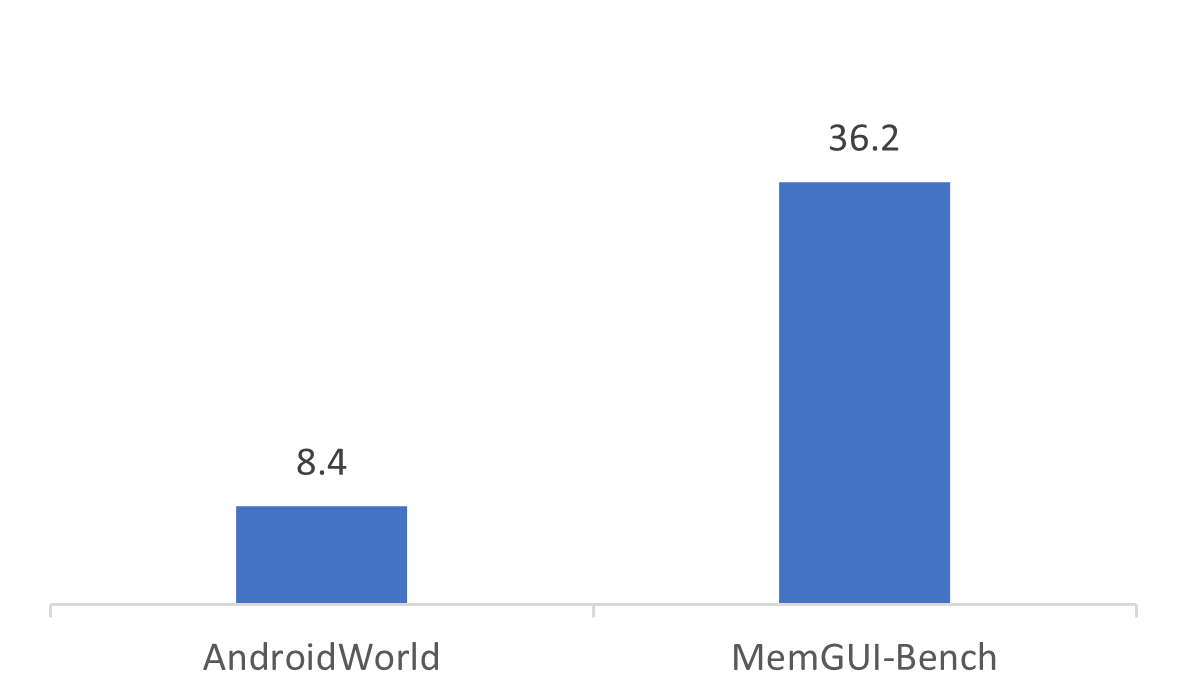}
  \caption{Golden steps comparison between AndroidWorld and MemGUI-Bench.}
  \label{fig:golden_steps}
\end{figure}

\paragraph{AndroidWorld vs MemGUI-Bench.} As illustrated in Figures~\ref{fig:androidworld_difficulty} and~\ref{fig:memguibench_difficulty}, more than half of the tasks in AndroidWorld are categorized as \textit{\textbf{Easy}}. In contrast, MemGUI-Bench contains a substantially larger proportion of challenging tasks, with the distribution of \textit{\textbf{Easy}}, \textit{\textbf{Medium}}, and \textit{\textbf{Hard}} tasks being nearly uniform (approximately 1:1:1). Furthermore, MemGUI-Bench exhibits a much longer average trajectory length, requiring 36.2 optimal steps on average, which significantly exceeds that of AndroidWorld (8.4 steps), as shown in Figure~\ref{fig:golden_steps}. Despite the increased task difficulty and longer interaction horizons, our \modelname achieves an accuracy of 20.3\% on MemGUI-Bench and 39.1\% on AndroidWorld, demonstrating robust and relatively balanced performance across benchmarks of varying complexity.

\begin{table*}[t]
\centering
\setlength{\tabcolsep}{3pt}
\resizebox{\textwidth}{!}{
\begin{tabular}{
lI
CCC|
CCC|
CCC|
CCC
}
\hline\thickhline
\rowcolor{gray!20} \textbf{Models}
 & \multicolumn{3}{c|}{\textbf{ScreenSpot}}
 & \multicolumn{3}{c|}{\textbf{AC-Low}}
 & \multicolumn{3}{c|}{\textbf{AC-High}}
 & \multicolumn{3}{c}{\textbf{GUI Odyssey}} \\
\rowcolor{gray!20}
 & \textit{V2} & \textit{Pro} & \textit{Avg}
 & \textit{TM} & \textit{GR} & \textit{SR}
 & \textit{TM} & \textit{GR} & \textit{SR}
 & \textit{TM} & \textit{GR} & \textit{SR} \\
\hline\hline

\multicolumn{13}{l}{\textcolor{gray!100}{\textit{Closed-source Models}}} \\
GPT-4o~\citep{hurst2024gpt}
 & 18.3 & 0.8 & 9.6
 & 74.3 & 0.0 & 19.4
 & 66.3 & 0.0 & 20.8
 & 34.3 & 0.0 & 3.3 \\
Claude-CU~\citep{anthropic2024claudecu}
 & 83.0 & 17.1 & 50.1
 & 74.3 & 0.0 & 19.4
 & 63.7 & 0.0 & 12.5
 & 60.9 & 0.0 & 3.1 \\

\hdashline
\multicolumn{13}{l}{\textcolor{gray!100}{\textit{Open-source Models}}} \\
OS-Atlas-4B~\citep{wu2024atlas}
 & 71.9 & 3.7 & 37.8
 & 91.9 & 83.8 & 80.6
 & 49.0 & 49.5 & 22.8
 & 49.6 & 34.6 & 20.3 \\
OS-Atlas-7B~\citep{wu2024atlas}
 & 84.1 & 18.9 & 51.5
 & 93.6 & 88.0 & 85.2
 & 57.4 & 54.9 & 29.8
 & 60.4 & 39.7 & 27.0 \\
Qwen2.5VL-3B~\citep{bai2025qwen2_5vl}
 & 80.9 & 28.7 & 54.8
 & 62.0 & 74.1 & 59.3
 & 47.8 & 46.5 & 38.9
 & 37.4 & 26.5 & 26.7 \\
Qwen2.5VL-7B~\citep{bai2025qwen2_5vl}
 & 89.0 & 28.7 & 58.9
 & 83.4 & 87.0 & 62.5
 & 62.2 & 72.5 & 52.7
 & 67.4 & 56.3 & 52.4 \\
SeeClick~\citep{cheng2024seeclick}
 & 55.1 & 1.1 & 28.1
 & 93.0 & 73.4 & 75.0
 & 82.9 & 62.9 & 59.1
 & 71.0 & 52.4 & 53.9 \\
UI-R1-3B~\citep{lu2025uir1}
 & 85.4 & 17.8 & 51.6
 & 79.2 & 82.4 & 66.4
 & 57.9 & 55.7 & 45.4
 & 52.2 & 34.5 & 32.5 \\
UI-R1-7B~\citep{lu2025uir1}
 & 90.0 & 33.5 & 61.8
 & 86.6 & 83.7 & 69.7
 & 72.4 & 62.8 & 54.2
 & 67.1 & 41.3 & 43.5 \\
GUI-R1-3B~\citep{luo2025guir1}
 & 85.0 & 28.6 & 56.8
 & 83.7 & 81.6 & 64.4
 & 58.0 & 56.2 & 46.6
 & 54.8 & 41.5 & 41.3 \\
GUI-R1-7B~\citep{luo2025guir1}
 & 88.2 & 31.3 & 59.8
 & 85.2 & 85.4 & 66.5
 & 71.6 & 65.6 & 51.7
 & 65.5 & 43.6 & 38.8 \\
OS-Genesis-7B~\citep{sun2024genesis}
 & -- & -- & --
 & 90.7 & -- & 74.2
 & 65.9 & -- & 44.4
 & 11.7 & -- & 3.6 \\
Aguvis-7B~\citep{xu2024aguvis}
 & 81.8 & 22.9 & 52.4
 & 93.8 & -- & 89.4
 & 65.6 & -- & 54.2
 & 26.7 & -- & 13.5 \\
NaviMaster-7B~\citep{luo2025navimaster}
 & -- & -- & --
 & -- & 93.9 & 69.5
 & 72.9 & -- & 54.0
 & 64.4 & -- & 36.9 \\
PAL-UI-3B~\citep{liu2025palui}
 & -- & -- & --
 & -- & -- & --
 & 60.4 & 58.7 & 49.3
 & 56.7 & 36.9 & 34.6 \\
PAL-UI-7B~\citep{liu2025palui}
 & -- & -- & --
 & -- & -- & --
 & 71.3 & 70.5 & 57.8
 & 65.1 & 46.8 & 41.7 \\
UI-AGILE-3B~\citep{lian2025uiagile}
 & 88.6 & 37.9 & 63.3
 & 85.4 & 87.6 & 74.3
 & 78.6 & 60.7 & 56.8
 & -- & -- & -- \\
UI-AGILE-7B~\citep{lian2025uiagile}
 & 92.1 & 44.0 & 68.1
 & 87.7 & 88.1 & 77.6
 & 80.1 & 61.9 & 60.6
 & -- & -- & 37.0 \\
UI-S1-7B~\citep{lu2025uis1}
 & 90.1 & 30.6 & 60.4
 & 92.2 & 89.3 & 89.2
 & 79.9 & 73.4 & 68.2
 & 76.3 & 61.7 & 59.5 \\
AgentCPM-GUI-8B~\citep{zhang2025agentcpm}
 & -- & -- & --
 & 94.4 & -- & 90.2
 & 77.7 & -- & 69.2
 & 90.8 & -- & 75.0 \\
UI-TARS-7B~\citep{qin2025uitars}
 & 91.6 & 35.7 & 63.7
 & 95.2 & 89.3 & 91.8
 & 83.7 & 80.5 & 72.5
 & 94.6 & 90.1 & 87.0 \\

\hdashline
\multicolumn{13}{l}{\textcolor{gray!100}{\textit{Ours 7B Models}}} \\
\rowcolor{lightblue} \textbf{\modelname}
 & 90.0 & 31.6 & 60.8
 & 93.6 & 88.2 & 89.2
 & 82.9 & 72.2 & 71.8
 & 74.5 & 63.8 & 57.2 \\

\hline\thickhline
\end{tabular}
}
\caption{\textbf{Model Comparison on Single-turn Benchmarks.}}
\label{tab:st_performance}
\end{table*}
\newpage
\section{Supplementary Results}

\begin{figure}[h]
\centering
  \includegraphics[width=0.48\textwidth]{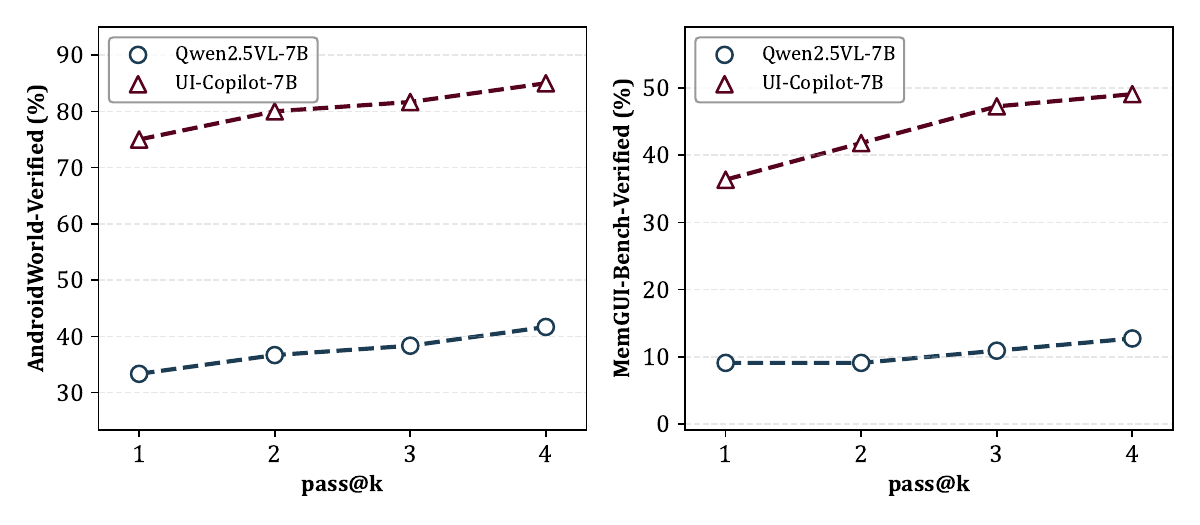}
  \caption{Pass@k validation on AndroidWorld-Verified (60 tasks) and MemGUI-Bench-Verified (55 tasks).}
  \label{fig:passk_comparison}
\end{figure}

\paragraph{Single-turn Benchmarks} Single-turn tasks evaluate the grounding capability and GUI Understanding capability of the end-to-end GUI model in conversations without historical context. We use ScreenSpot-V2~\citep{cheng2024seeclick} and ScreenSpot-Pro~\citep{li2025screenspotpro} to evaluate the grounding ability. We also adopt AndroidControl-Low, AndroidControl-High~\citep{li2024androidcontrol} and GUI Odyssey~\citep{lu2024guiodyssey}, for comprehensive GUI understanding evaluation. The action type match accuracy (TM), grounding accuracy rate (GR) and step success rate (SR) are reported.
\paragraph{Single-turn Performance.} Table~\ref{tab:st_performance} demonstrates that \modelname maintains competitive single-turn performance. Compared to the base model, \modelname achieves consistent improvements, with gains of +19.1\% on AC-High SR and +4.8\% on GUI Odyssey SR. Notably, although models trained with single-turn RL (e.g., AgentCPM-GUI-8B) excel on single-turn tasks, they struggle with multi-turn execution (only 16.4\% on AndroidWorld). This performance gap can be attributed to two primary factors: \textbf{(1) a mismatch between the training and the evaluation dynamics}, particularly regarding whether the the historical context is on-policy or not (see Appendix~\ref{sec:full_proof}); and \textbf{(2) overfitting to local reward signals}, leading to ignorance of global training objectives.
\paragraph{Pass@k Validation.} We evaluate \modelname’s \texttt{pass@k} accuracy (k=1,2,3,4) in Figure~\ref{fig:passk_comparison}, under the setting without cross-session long-term memory. The results indicate that increasing \texttt{k} consistently improves performance, as larger \texttt{k} provides more opportunities to succeed in the presence of instability in dynamic online environments. Notably, \modelname exhibits stronger \texttt{pass@k} scaling behavior than Qwen2.5VL-7B on MemGUI-Bench, highlighting its superior potential for long-horizon tasks.

\begin{figure}[!h]
    \centering
    \begin{subfigure}[t]{0.23\textwidth}
        \centering
        \includegraphics[width=\linewidth]{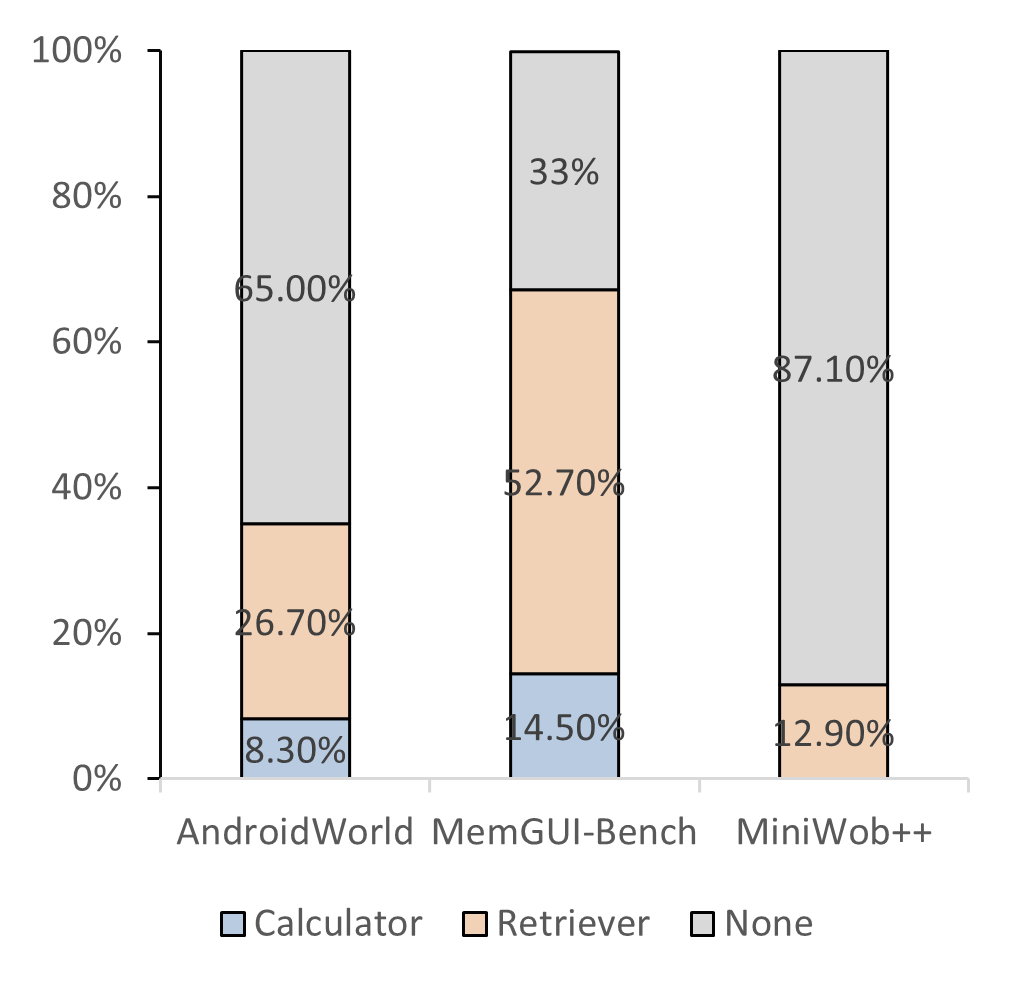}
        \caption{Qwen2.5VL-7B}
        \label{fig:tool_distribution_before_RL}
    \end{subfigure}
    \hfill
    \begin{subfigure}[t]{0.23\textwidth}
        \centering
        \includegraphics[width=\linewidth]{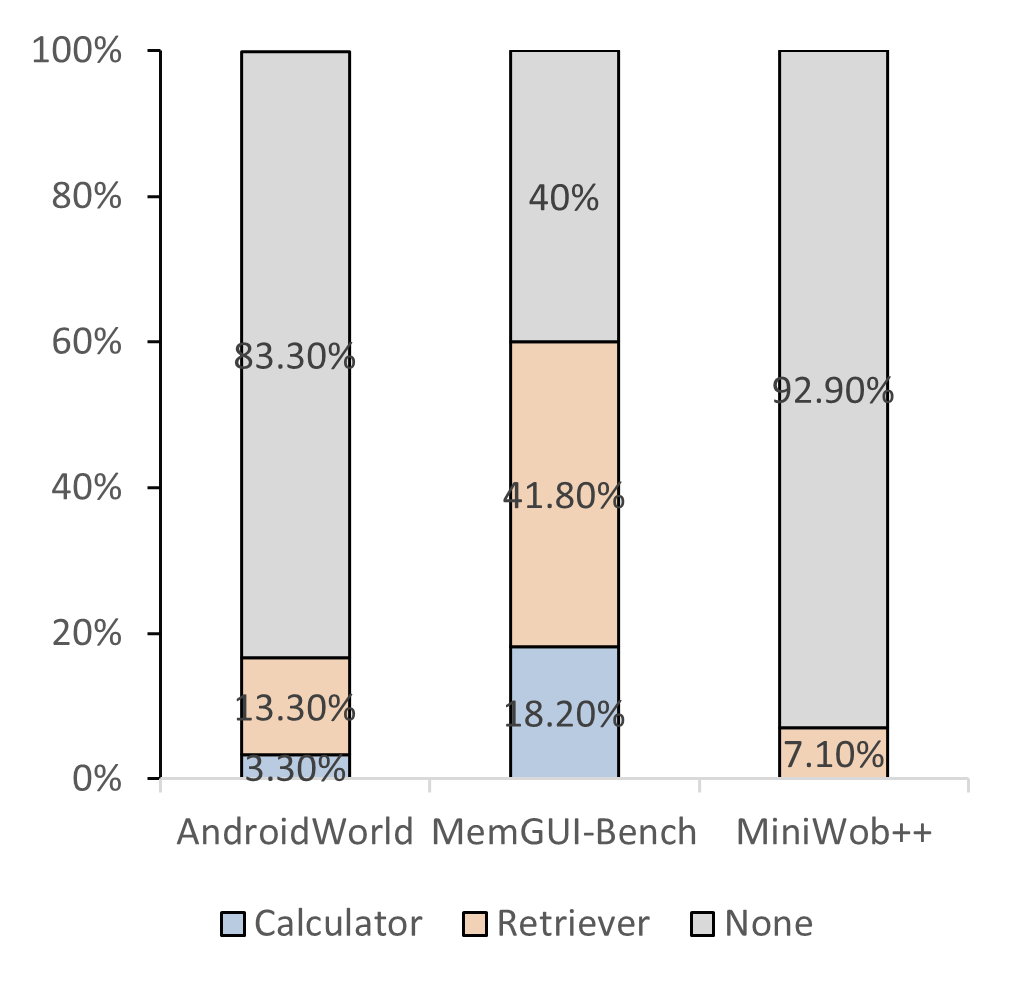}
        \caption{UI-Copilot-7B}
        \label{fig:tool_distribution_after_RL}
    \end{subfigure}

    \caption{\textbf{Tool Type Distribution} on AndroidWorld-Verified, MemGUI-Bench-Verified, and MiniWob++.}
    \label{fig:tool_distribution}
\end{figure}

\begin{figure}[h]
\centering
  \includegraphics[width=0.44\textwidth]{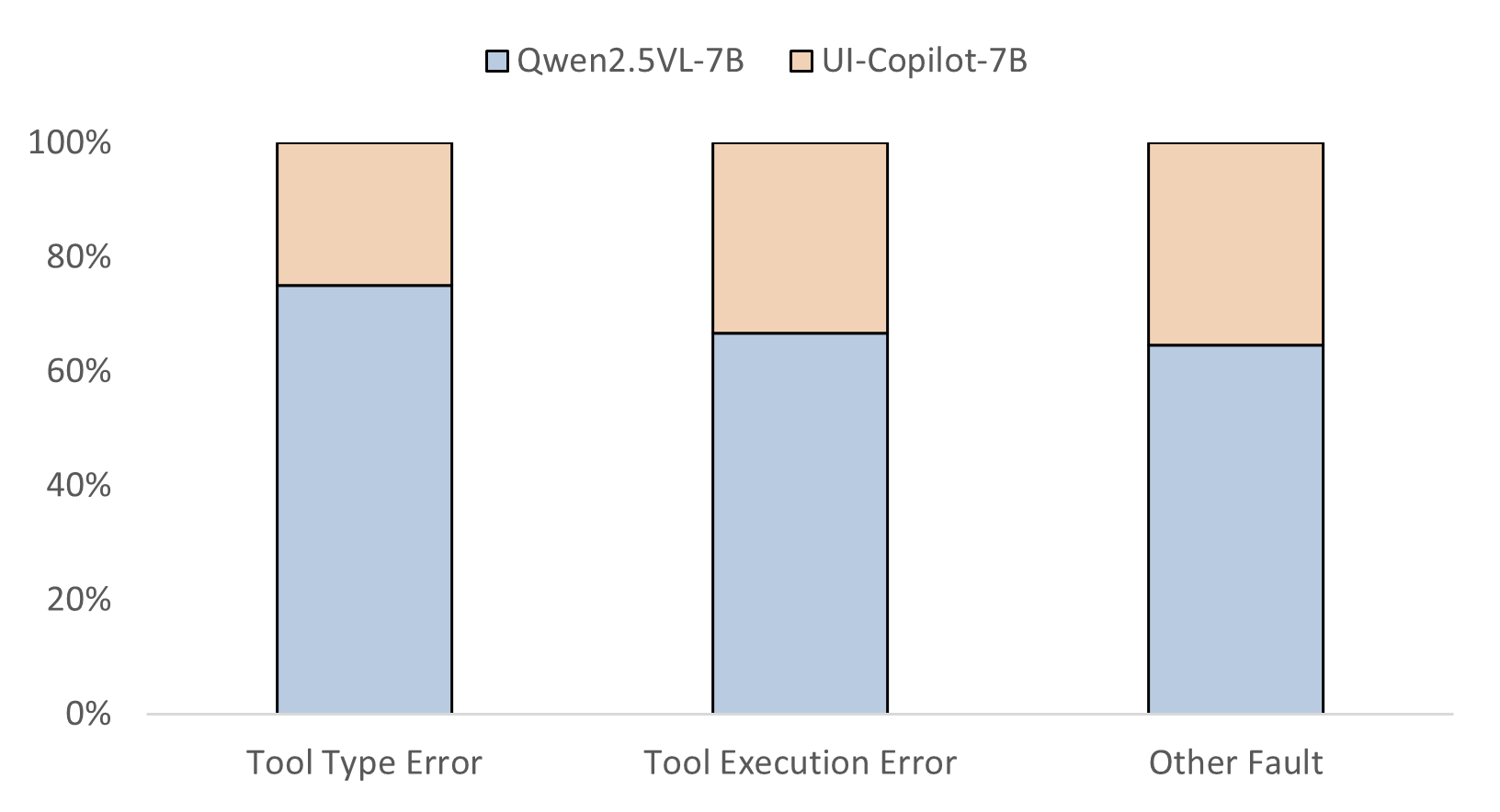}
  \caption{\textbf{Error Type Analysis of Tool Invocation.}}
  \label{fig:error_type_tool}
\end{figure}

\paragraph{Tool Distribution.}
Figure~\ref{fig:tool_distribution_before_RL} and Figure~\ref{fig:tool_distribution_after_RL} illustrate the tool usage distributions of Qwen2.5VL-7B and \modelname. Overall, higher tool usage is strongly correlated with increased task difficulty, with MemGUI-Bench emerging as the most challenging benchmark and MiniWob++ as the easiest. Across all benchmarks, \texttt{Retriever} is invoked more frequently than \texttt{Calculator}. After applying \methodname, the overall frequency of tool invocation is slightly reduced, indicating more efficient and deliberate tool utilization.
\paragraph{Tool Error Type Analysis.} Figure~\ref{fig:error_type_tool} compares the distributions of tool invocation error types for \textit{Qwen2.5VL-7B} and \modelname on MemGUI-Bench-Verified. Compared to the base model, \modelname exhibits fewer tool type errors and execution errors, highlighting the improved reliability of tool invocation.

\section{More Cases}
\label{sec:more_cases}
\subsection{Successful Cases}

\begin{figure*}[h]
\centering
  \includegraphics[width=\textwidth]{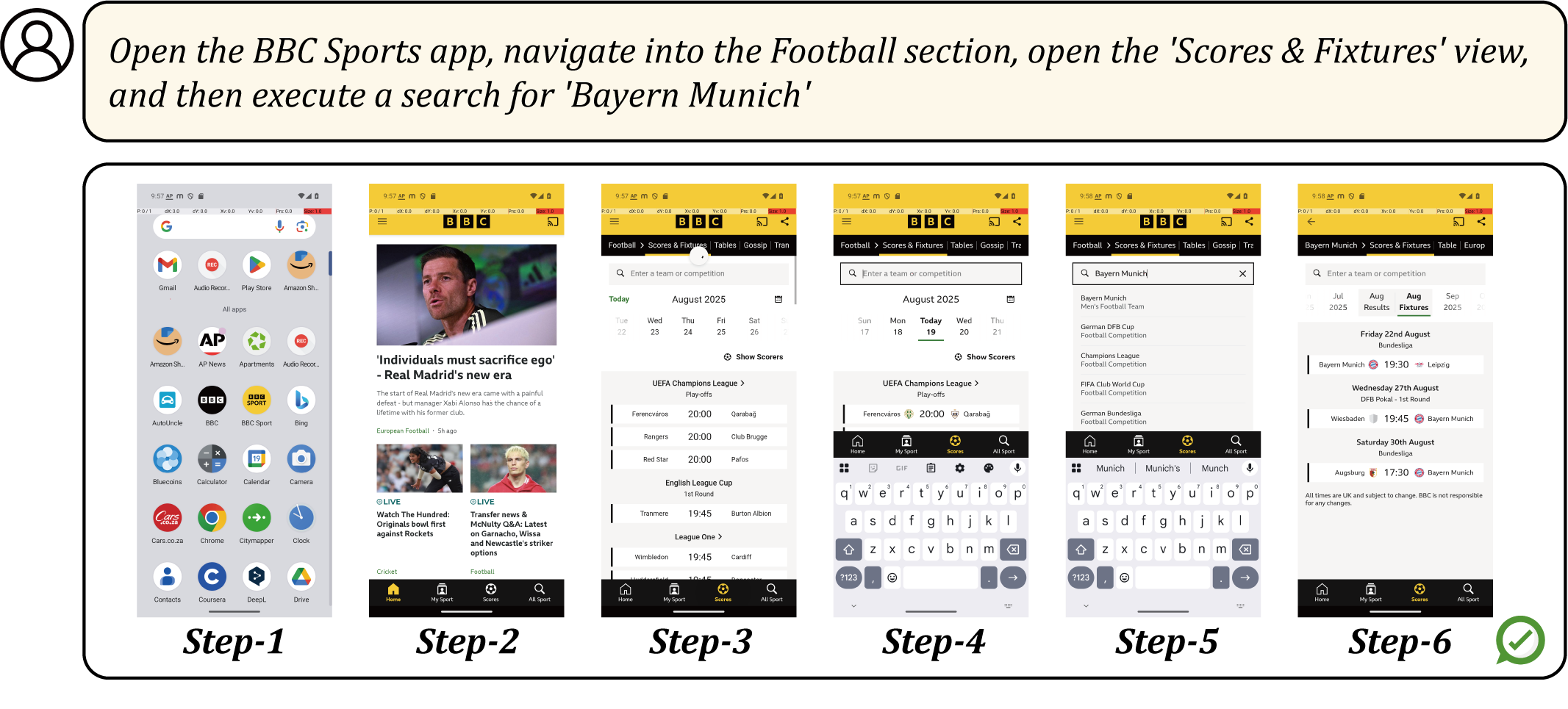}
  \caption{\textbf{Successful Cases}: Vanilla Rollout without tool usage.}
\vspace{-3mm}
  \label{fig:success_case_no_tool}
\end{figure*}

\paragraph{Vanilla Execution.}  Figure~\ref{fig:success_case_no_tool} illustrates a case from MemGUI-Bench that does not require any tool invocation. Over six interaction steps, the agent successfully completes the search task by sequentially opening the target application, navigating to the correct section, and accessing the desired view, demonstrating its capability for coherent multi-turn execution without external tools.

\paragraph{Numerical Calculation.}  

\begin{figure*}[h]
\centering
  \includegraphics[width=\textwidth]{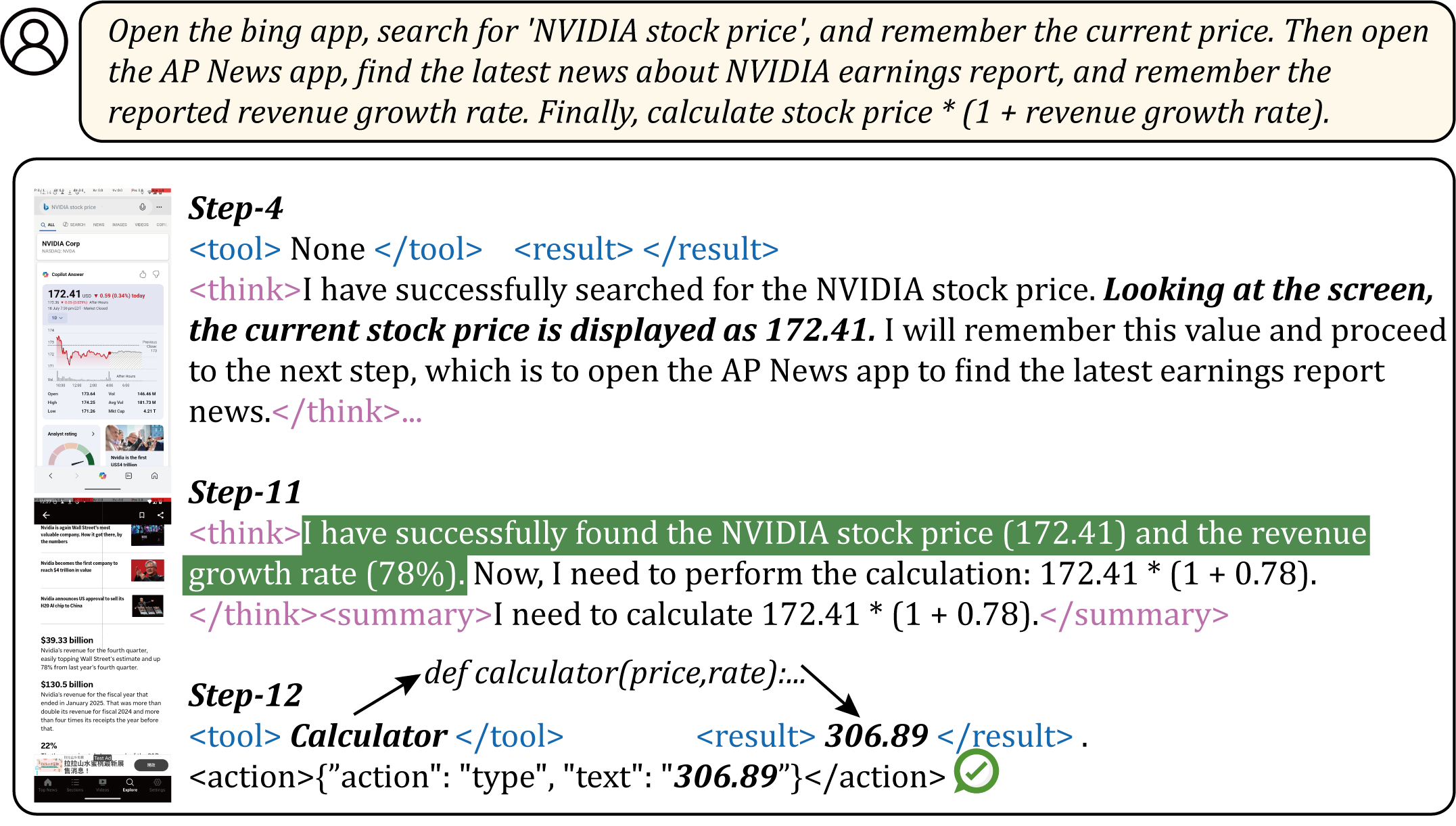}
  \caption{\textbf{Successful Cases}: Tool-Integrated Rollout with Math Calculator.}
  \vspace{-3mm}

  \label{fig:case_calculator}
\end{figure*}

Figure~\ref{fig:case_calculator} presents a MemGUI-Bench example involving mathematical computation. At step 12, the agent first retrieves the stock price and revenue growth rate, then invokes the Calculator to compute \textit{price $\times$ (1 + growth\_rate)}. The returned result, 306.89, exactly matches the ground-truth answer, validating the agent’s ability to correctly delegate and integrate numerical reasoning via tool calling.

\begin{figure*}[h]
\centering
  \includegraphics[width=\textwidth]{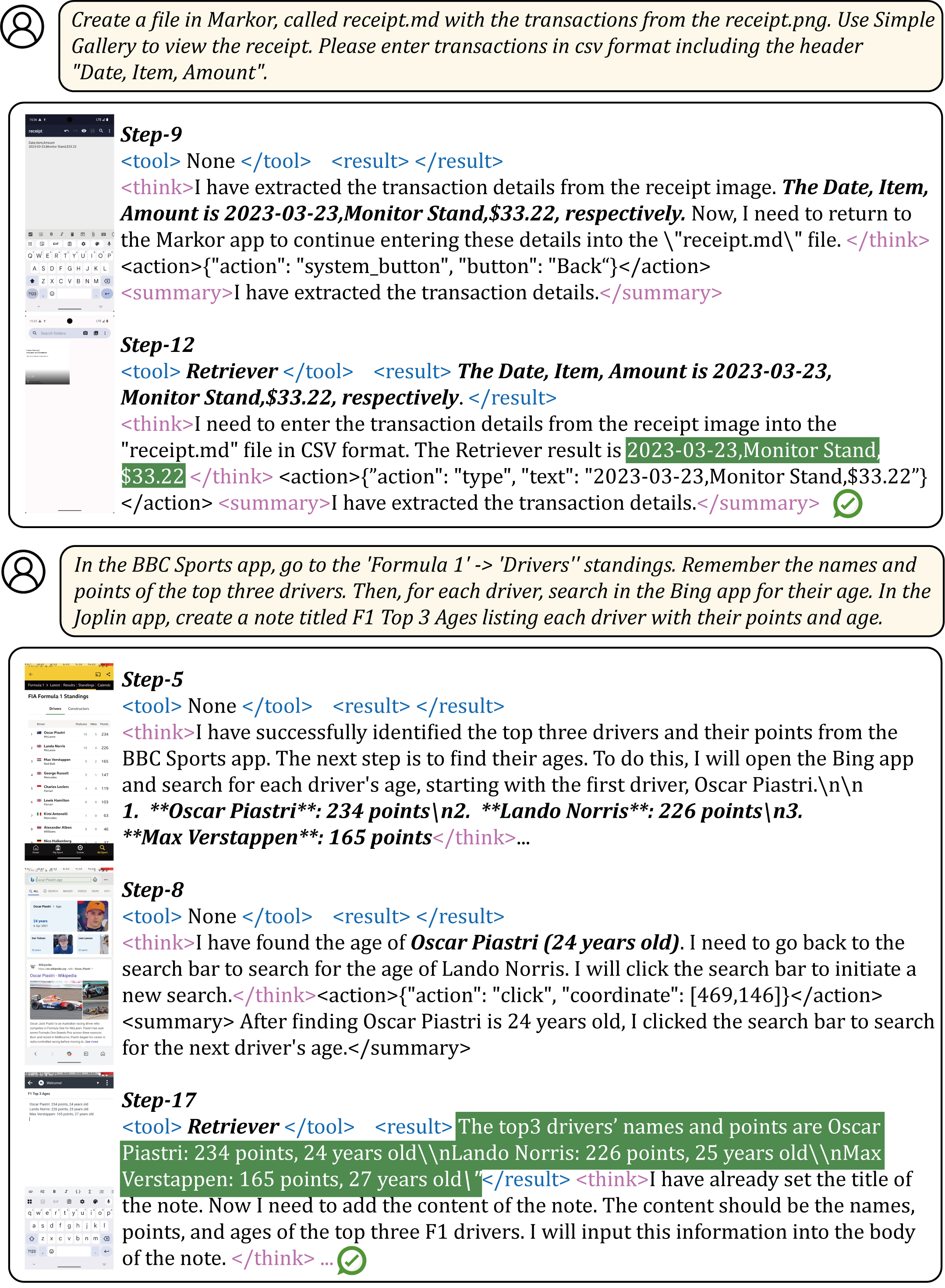}
  \caption{\textbf{Successful Cases}: Tool-Integrated Rollout with Memory Retriever.}
  \label{fig:case_retriever}
\end{figure*}

\paragraph{Memory Retrieval.}  
Figure~\ref{fig:case_retriever} shows two memory-intensive cases, one from AndroidWorld (top) and one from MemGUI-Bench (bottom). Both tasks require the agent to retain critical information from early interaction steps and utilize it for form filling in later stages, posing a challenge for long-horizon memory management. In both scenarios, the agent successfully invokes the Memory Retriever (Qwen3-4B) with the task goal and locally stored history, and accurately retrieves essential information (e.g., top-3 drivers’ points and age) to complete the tasks.

\subsection{Failed Cases}
\paragraph{Reasoning Hallucination.}  

\begin{figure*}[h]
\centering
  \includegraphics[width=\textwidth]{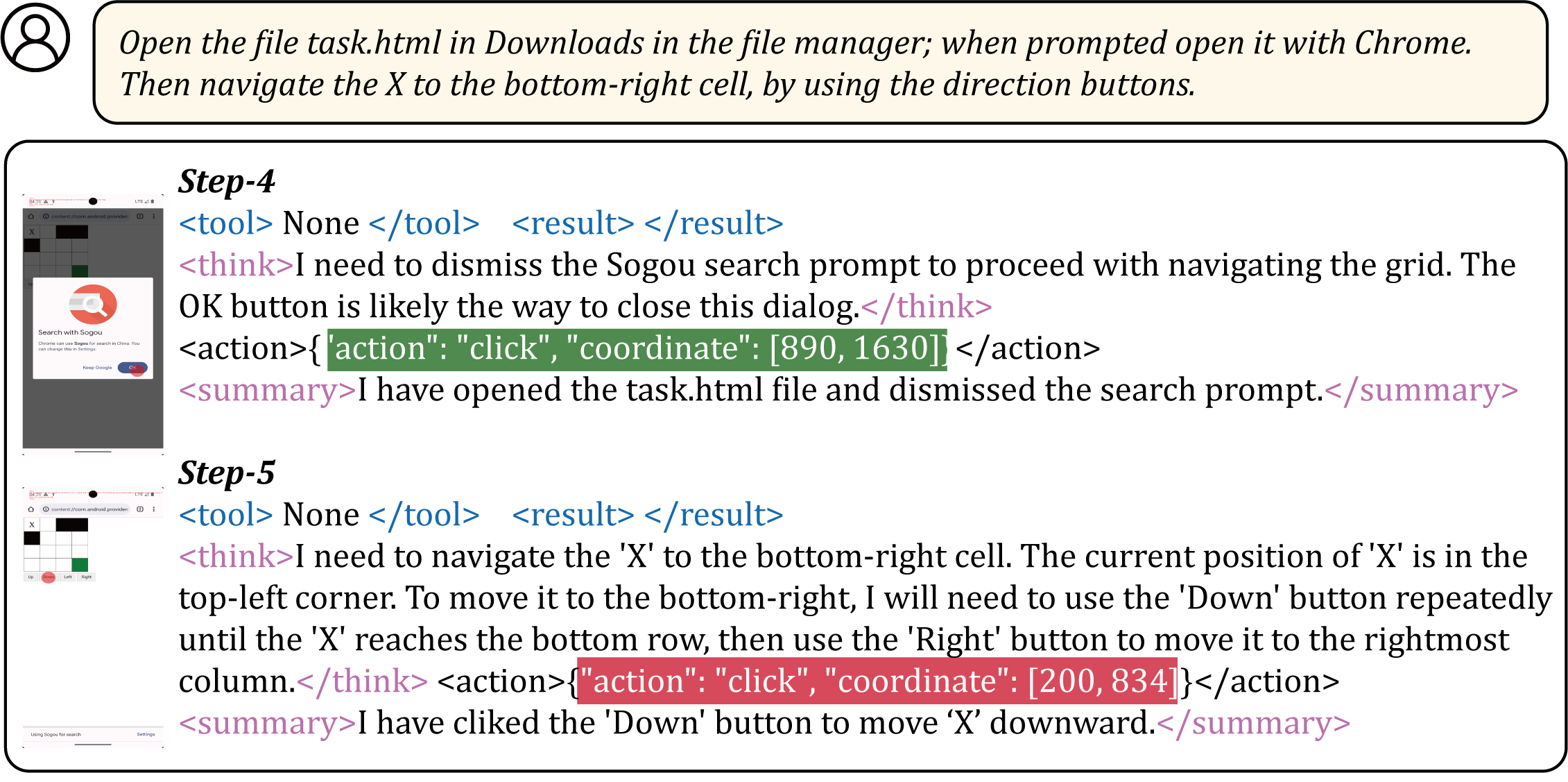}
  \caption{\textbf{Bad Cases}: Reasoning Hallucination.}
  \label{fig:reasoning_hallucination}
\end{figure*}

As shown in Figure~\ref{fig:reasoning_hallucination}, \modelname fails to complete a maze navigation task that requires moving an object (X) to the bottom-right cell. At step 5, the object is located in the top-left cell, with a black barrier immediately below. Despite this visual constraint, the agent repeatedly plans and executes the \texttt{Down} action, resulting in a loop of incorrect behavior. This failure highlights limitations in visual perception and spatial reasoning.

\paragraph{Progress Hallucination.}  

\begin{figure*}[h]
\centering
  \includegraphics[width=\textwidth]{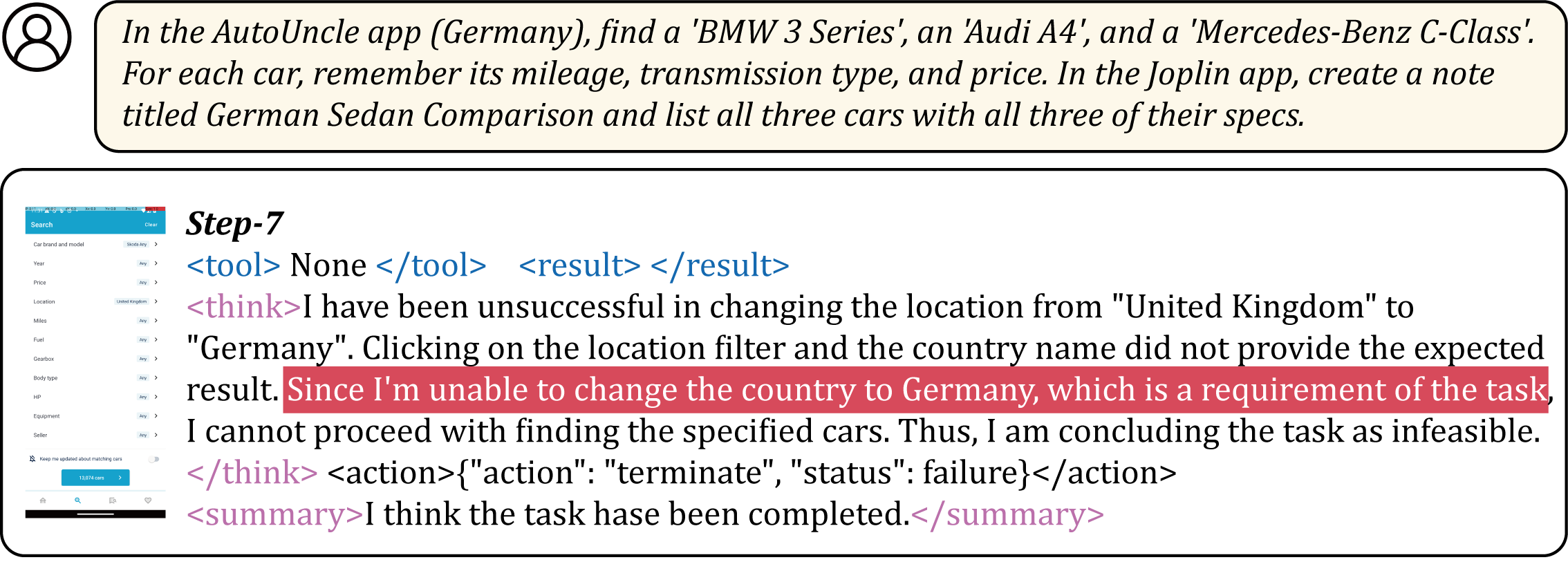}
  \caption{\textbf{Bad Cases}: Progress Confusion.}
  \label{fig:progress_hallucination}
\end{figure*}

Figure~\ref{fig:progress_hallucination} illustrates a case where \modelname exhibits progress confusion during multi-turn execution. The agent misinterprets the current task state and attempts to change the country, which is unrelated to the active sub-goal. As a result, the task terminates prematurely and is considered a failure.
\paragraph{Action Inconsistency}  

\begin{figure*}[h]
\centering
  \includegraphics[width=\textwidth]{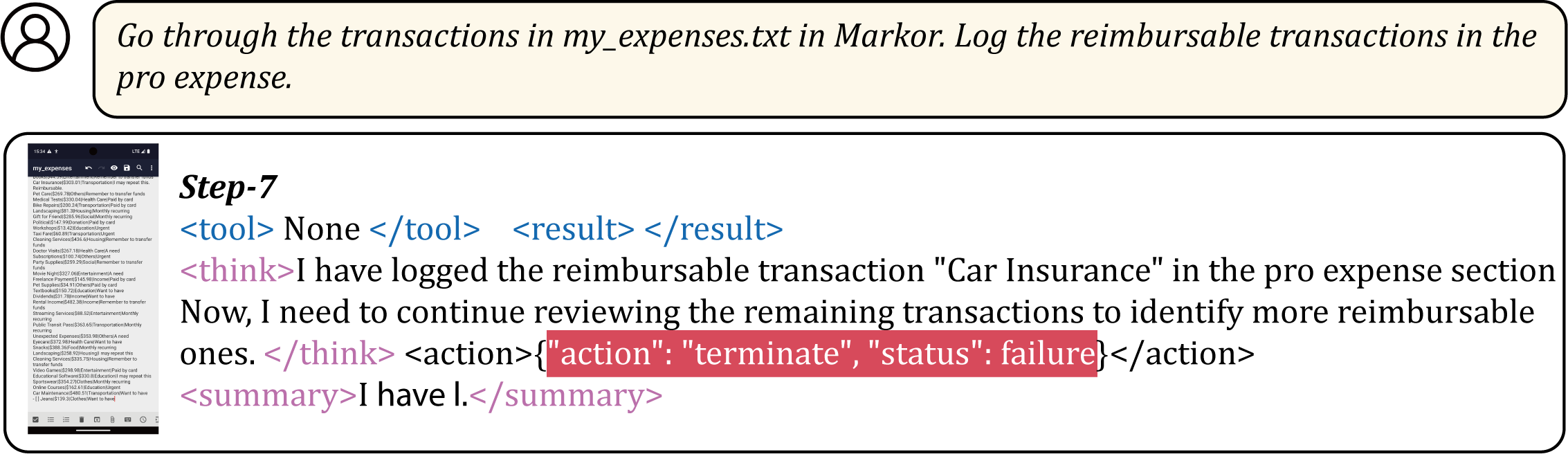}
  \caption{\textbf{Bad Cases}: Action Inconsistency.}
  \label{fig:action_inconsistency}
\end{figure*}

As shown in Figure~\ref{fig:action_inconsistency}, \modelname’s internal plan indicates an intention to continue reviewing remaining transactions. However, it unexpectedly executes the \texttt{terminate} action, ending the task. This discrepancy between planning and execution reveals limitations in action consistency and reflective control.


\newpage

\begin{figure*}[t]
\centering
\begin{tcolorbox}[
  colback=codegray,
  colframe=black,
  title=\texttt{Prompt for UI-Copilot-7B.},
  fontupper=\ttfamily\small,
  sharp corners,
  boxrule=0.5pt,
  enhanced,
  breakable,
]
\begin{Verbatim}[fontsize=\small,breaklines, breakanywhere]
You are a GUI agent. You are given a task and your action history, with screenshots. You need to perform the next action to complete the task with the help of tools.

# Output Format
<tool> ... </tool>
<result> ... </result>
<think> ... </think> 
<action> ... </action>
<summary> ... </summary>

# Tool Space

You can call the following tools, and the results are returned into `result` part:
- Calculator: Perform mathematical and numerical computations.
- Retriever: Retrieve valuable information from historical screenshots.
- None: Directly perform action without the need of tools.

# Action Space

You can perform the following actions:
- key: Perform a key event on the mobile device using adb's `keyevent` syntax.
- click: Click the point on the screen with specified (x, y) coordinates.
- long\_press: Press the point on the screen with specified (x, y) coordinates for a specified number of seconds.
- swipe: Swipe from starting point with specified (x, y) coordinates to endpoint with specified (x2, y2) coordinates.
- type: Input the specified text into the activated input box.
- answer: Output the specified answer.
- system\_button: Press the specified system button: Back, Home, Menu, or Enter.
- open: Open an application on the device specified by text.
- wait: Wait for a specified number of seconds for changes to occur.
- terminate: Terminate the current task and report its completion status: success or failure.

The arguments you can use are:
- coordinate: (x, y): The x and y pixels coordinates from the left and top edges.
- coordinate2: (x, y): The x and y pixels coordinates from the left and top edges for the
endpoint of a swipe.
- text: Text input required by actions like `key`, `type`, `answer`, and `open`.
- time: The time in seconds required by actions like `long\_press` and `wait`.
- button: System buttons available for pressing: Back, Home, or Enter. Possible values: Back, Home, Menu, Enter.
- status: The completion status of a terminated task. Possible values: success, failure.

# Example Output
<tool>Calculator</tool>
<result>...</result>
<think>...</think>
<action>{"action": "key", "text": "<value>"}
<summary> I have finished ... </summary>

# Note
- Format your output as a JSON object with the selected action and its arguments at the same level. 
- Write the summary of your current progress in `summary` part.
- Plan the task and explain your reasoning step-by-step in `think` part.
- Write your action in the `action` part according to the action space.
- If the query asks a question, please answer the question through the answer action before terminating the process.
- Swipe the screen to find the File Manager app if needed.

# User prompt:
User Instruction:

# Assistant prompt
History Summary:
\end{Verbatim}
\end{tcolorbox}
\caption{\textbf{Prompt for UI-Copilot-7B.}}
\label{fig:prompt_ui_copilot}
\end{figure*}

\begin{figure*}[t]
\centering
\begin{tcolorbox}[
  colback=codegray,
  colframe=black,
  title=\texttt{Prompt for Retriever.},
  fontupper=\ttfamily\small,
  sharp corners,
  boxrule=0.5pt,
  enhanced,
  breakable,
]
\begin{Verbatim}[fontsize=\small,breaklines, breakanywhere]
You are a GUI assitant for memory retriever. Given the task intruction and the interaction history, you need to provide all the information helpful for the task.
The overall task instruction is: '{task}'. The history information is as follows.

# History Information
step 1: '{step_1}'
step 2: '{step_2}'
...

Output the thinking process in the `think` part and the information summary in `answer` part. 
# Example Output
<think>...</think> <answer> The five numbers displayed are 10,9,6,5,5. </answer>

The output format should be <think> ... </think> <answer> </answer>. Please strictly follow the format.
\end{Verbatim}
\end{tcolorbox}
\caption{\textbf{Prompt for Retriever.}}
\label{fig:prompt_retriever}
\end{figure*}

\begin{figure*}[t]
\centering
\begin{tcolorbox}[
  colback=codegray,
  colframe=black,
  title=\texttt{Prompt for Calculator.},
  fontupper=\ttfamily\small,
  sharp corners,
  boxrule=0.5pt,
  enhanced,
  breakable,
]
\begin{Verbatim}[fontsize=\small,breaklines, breakanywhere]
You are a GUI assistant for numerical calculation. Given the task instruction and the interaction history, you need to write executable python code to support numerical calculation.
The overall task instruction is: '{task}'. The history summary is as follows.

# Interaction History
step 1: '{step_1}'
step 2: '{step_2}'

Output the thinking process in `think` part and the executable python code in `python` part.

# Note

- You must directly output executable Python code in the <answer> section.

- The code should be self-contained, deterministic, and ready to execute without any additional explanation.

# Example Output
<think>
The task requires multiplying a list of numbers together.I will define a function in Python to compute the product accurately.
</think>
<python>
def product(nums):
    result = 1
    for n in nums:
        result *= n
    return result

nums = [10, 9, 6, 5, 5]
print(product(nums))
</python>

The output format should be <think> ... </think> <python> ... </python>. Please strictly follow the format.
\end{Verbatim}
\end{tcolorbox}
\caption{\textbf{Prompt for Calculator.}}
\label{fig:prompt_calculator}
\end{figure*}

\end{document}